\documentclass{article} 
\usepackage{collas2026_conference,times}
\usepackage{easyReview}

\usepackage[utf8]{inputenc}
\usepackage{graphicx}
\usepackage{amsmath}
\usepackage{amssymb}
\usepackage{hyperref}
\usepackage{url}
\usepackage{float}
\usepackage{indentfirst}
\usepackage{booktabs}
\usepackage{enumitem}
\usepackage{subcaption}
\usepackage{wrapfig}
\usepackage{algorithm}
\usepackage{algorithmic}
\usepackage{dsfont}
\usepackage{cleveref}

\usepackage{array}     
\usepackage{comment}
\usepackage{xcolor}
\usepackage{etoc}
\usepackage{multirow}
\usepackage{wrapfig}
\usepackage{booktabs}
\usepackage{caption}
\usepackage{makecell}

\newcommand{\ci}[1]{{\scriptstyle \pm #1}}

\usepackage{amsmath,amsfonts,bm}

\def\eqref#1{equation~\ref{#1}}

\def\1{\bm{1}}

\DeclareMathAlphabet{\mathsfit}{\encodingdefault}{\sfdefault}{m}{sl}
\SetMathAlphabet{\mathsfit}{bold}{\encodingdefault}{\sfdefault}{bx}{n}

\usepackage{hyperref}
\hypersetup{
    colorlinks=true,
    linkcolor=black,    
    filecolor=magenta,
    urlcolor=blue,
    citecolor=purple,
    pdftitle={Overleaf Example},
    pdfpagemode=FullScreen,
    }

\title{Benchmarking Unlearning for Vision Transformers}

\author{Kairan Zhao\thanks{Equal contribution} \ \thanks{Correspondence
to: Kairan.Zhao@warwick.ac.uk}\\ 
 University of Warwick\\
 \And
 Iurie Luca\footnotemark[1]\\ 
 University of Warwick\\
\And
Peter Triantafillou\\
 University of Warwick\\
}

\collasfinalcopy 

\begin{document}

\maketitle

\begin{abstract}
Machine unlearning (MU) refers to the post-training model capability to adapt by removing (the influence of) training examples that are incorrect, biased, or leaking sensitive/private information.  
MU is now widely regarded as a critical capability for building safe and fair AI. 
In parallel, research into transformer architectures for computer vision tasks has been highly successful: Increasingly, Vision Transformers (VTs) emerge as strong alternatives to CNNs. 
Yet, MU research for vision tasks has largely centered on CNNs, not VTs. 
While MU benchmarks have been developed for LLMs, diffusion models, and CNNs, none currently exist for VTs.
\emph{This work is the first to attempt this, benchmarking MU algorithm performance across different VT families (ViT, Swin-T, and DINOv2) and at different capacities}.
The work employs (i) different datasets, selected to assess the impacts of dataset scale and complexity; 
(ii) different MU algorithms, selected to represent fundamentally different approaches for MU; and (iii) both single-shot and continual unlearning protocols. 
Additionally, it focuses on benchmarking MU algorithms that leverage training data memorization, since leveraging memorization has been recently discovered to significantly improve the performance of previously SOTA algorithms. En route, the work characterizes how VTs memorize training data relative to CNNs, and assesses the impact of different memorization proxies on performance. 
The benchmark uses unified evaluation metrics that capture two complementary notions of forget quality along with accuracy on unseen (test) data and on retained data. Overall, this work offers a benchmarking basis, enabling reproducible, fair, and comprehensive comparisons of existing (and future) MU algorithms on VTs. 
Importantly, for the first time, it sheds light on how well existing algorithms work in VT settings, establishing a promising reference performance baseline. 
\end{abstract}

\section{Introduction}\label{intro}
The success of Vision Transformers (VTs) brings with it new responsibilities regarding responsible, fair, and safe AI. Among these, the ability to remove (the influence of) specific ``problematic'' data from trained models (a.k.a. machine unlearning) (MU) is critical. 
Said problematic data may include biased, erroneous, poisoned, obsolete, or privacy-sensitive data. Popular  VT architectures 
therefore represent an important frontier in this challenge.
In parallel, memorization has recently been identified as playing a fundamental role for MU (and this holds across modalities). 
For instance, MU research in LLMs \citep{barbulescu2024textualsequenceownimproving,jang2022knowledgeunlearningmitigatingprivacy} and in diffusion models \citep{eccv24mm,iclr24mm} explicitly detects and mitigates memorization. In computer-vision tasks, memorization has been shown to be a key factor affecting unlearning performance \citep{zhao2024makesunlearninghard,torkzadehmahani2024improved}.

Despite the fact that memorization and MU have been extensively studied in LLMs, text-to-image Diffusion Models, and Convolutional Networks (CNNs), it is an open question whether findings transfer to VTs. This uncertainty stems from key differences in architecture, training regimes, and inductive biases. 
Compared to LLMs (with which they share a transformer backbone), VTs operate on image patches using global self-attention (unlike LLMs which rely on causal masking or token-order constraints and process semantically meaningful tokens) and VTs are typically (pre)trained using supervised classification objectives (e.g., on ImageNet), while LLMs are trained on large corpora using self-supervised objectives, yielding different types of data exposure and memorization. 
Compared to CNNs, VTs lack strong spatial inductive biases such as locality (semantically related neighbouring pixels) and translation equivariance (for positional reasoning). These biases help CNNs localize and isolate memorization effects, whereas VTs must learn such structure from data alone, making VTs more data-hungry, often necessitating a pretrain–then–finetune training regime.
Moreover, unlike LLMs, VTs lack the benefit of language's syntactic and semantic structure. Hence, learned representations are more entangled and spread across layers and attention heads.

\textbf{The Gap.} One can thus reasonably expect the above differences to collectively introduce unique challenges for unlearning in VTs. Said potential challenges remain largely unexplored, despite a few related research works on both the algorithmic side (e.g., \citet{cadet2024deepunlearnbenchmarkingmachine,cho2024vitmulbaselinestudyrecent} whose evaluations of CNN-derived unlearning also include some VTs, typically, ViT-Tiny) and on the benchmarking side, with 
recent efforts systematically evaluating MU across tasks and modalities (such as 
\citet{maini2024tofutaskfictitiousunlearning,li2024wmdpbenchmarkmeasuringreducing} for LLMs, 
\citet{ma2024datasetbenchmarkcopyrightinfringement,zhang2024unlearncanvasstylizedimagedataset}, 
and \citet{cheng2024mubenchmultitaskmultimodalbenchmark} for text-to-image diffusion models). 

Focusing on image-classification tasks in computer vision,
the first comprehensive attempt in this domain, by \citet{triantafillou2024makingprogressunlearningfindings}, was based on the NeurIPS 2023 MU competition, evaluating and ranking the top algorithms on two datasets from the competition on a CNN (ResNet-18) architecture.
Interestingly, they report that MU methods can be brittle across architectures/datasets.
\citet{grimes2024goneforgottenimprovedbenchmarks} benchmarked MU on the same CNN environment but focused on evaluating more demanding privacy threats and paid attention to continual MU. 
\citet{cadet2024deepunlearnbenchmarkingmachine} extended the benchmark in \citet{triantafillou2024makingprogressunlearningfindings}, evaluating 18 MU algorithms on four datasets on a ResNet (as well as on ViT-Tiny).
Each of the above benchmarks has a different emphasis: Each is typically designed for a specific architecture-modality-task pairing. 
Some stress new datasets, while others stress comprehensive/exhaustive examinations/rankings of a large set of MU algorithms. 

\textbf{This work.} 
Overall, benchmarking for MU has been addressed for LLMs, diffusion models, and CNNs, but not for VTs.
We fill this gap by benchmarking MU on VTs, contributing an evaluation basis along the following key axes:
\begin{enumerate}
\item \textbf{Memorization:} Do VTs memorize differently from CNNs? How does this affect unlearning?
\item \textbf{Proxies:} Are CNN-based memorization proxies effective in VTs? Can they improve unlearning?
\item \textbf{Algorithms:} How do fundamentally different approaches for CNN-based MU perform on VTs?
\item \textbf{VT architectures:} How do VT design choices (ViT vs.\ Swin-T vs. DINOv2) and model capacity affect unlearning?
\item \textbf{Pretrain–Finetune:} How does the VT's pretrain–finetune paradigm influence MU?
\item \textbf{Algorithm-Architecture Pairings:} Are certain pairings especially compatible?
\item \textbf{Continual MU on VTs:} How stable is performance under continual unlearning?
\end{enumerate}

In contrast to prior ``standard'' benchmarking studies that are leaderboard-style (benchmarking/ranking a large set of algorithms) or proposing new datasets, our work follows a different path: It is centered on VTs, systematically testing \emph{representative} 
MU approaches across \emph{three} VT families (ViT, Swin-T, DINO) at \emph{different} capacities, over \emph{four} datasets of varying size/complexity, and under \emph{both} single-shot and continual unlearning scenarios. 
We employ unified metrics (ToW, ToW-MIA) that jointly account for retain/test accuracy, forget accuracy, and MIA vulnerability. We contrast results against CNN-derived MU counterpart algorithms. 
We aim to isolate architecture, capacity, memorization (and proxy) effects to assess how well MU methods perform in VTs and establish a strong performance baseline for MU in VTs.
The code for reproducing the results is available at:
\url{https://github.com/kairanzhao/Unlearning_VTs}

\section{Model Architectures and Algorithms}\label{related_work}

\subsection{Vision Transformers}
Transformer architectures \citep{VaswaniSPUJGKP17} leverage self-attention mechanisms instead of recurrence. 
We focus on, arguably, the most popular VTs: ViT, Swin-T, and DINOv2.
ViT \citep{dosovitskiy2021imageworth16x16words} processes images as sequences of flattened, embedded patches, which are (i) projected into an embedding space, (ii) combined with positional encodings, and (iii) passed through transformer encoder layers comprising multi-headed self-attention (MSA) and multi-layer perceptrons (MLP). 
Swin-T \citep{liu2021swintransformerhierarchicalvision} 
adds hierarchical representations (à la CNNs) by progressively merging patches and reducing spatial resolution at deeper layers, forming multi-scale representations. 
Additionally, a shifted-window self-attention mechanism is introduced to model both local and global contexts.
DINOv2 \citep{oquab2023dinov2} is a self-supervised ViT-based model trained on large-scale curated data, producing robust and transferable visual representations.
These three model families allow us to study MU performance across architectures which, due to having/lacking hierarchical representations or supervised/self-supervised pretraining, are more/less similar to CNNs (Swin-T vs. ViT/DINOv2).

\subsection{Machine Unlearning (MU)} \label{sec:mu}
MU \citep{cao2015machine_unlearning} aims to remove the influence of specific (``problematic'') training examples from pretrained models. A large body of work has tackled MU, spanning 
formal definitions and guarantees
\citep{ginart2019makingaiforgetyou,sekhari2021rememberwantforgetalgorithms,neel2020descenttodeletegradientbasedmethodsmachine}, 
empirically effective methods
\citep{goel, golatkar2020eternalsunshinespotlessnet,thudi2022unrollingsgdunderstandingfactors}, and other earlier efforts \citep{xu2023machineunlearningsurvey}. 
Over time, several key baselines have emerged, capturing fundamental components for successful MU. 
These range from simpler baselines like Fine-tune \citep{warnecke2023machineunlearningfeatureslabels, golatkar2020eternalsunshinespotlessnet}, to more sophisticated baselines like NegGrad+ \citep{kurmanji2023unboundedmachineunlearning}. 
Most recent SOTA algorithms include SCRUB \citep{kurmanji2023unboundedmachineunlearning}, 
L1-sparse \citep{jia2024modelsparsitysimplifymachine}, 
SalUn \citep{fan2024salunempoweringmachineunlearning} and the 
meta-algorithmic framework RUM \citep{zhao2024makesunlearninghard}. 

Evidently, all of the above methods have focused on CNNs. 
Our work aims to answer whether such 
CNN-derived MU algorithms transfer to VTs across VT families, model capacities, datasets, and protocols. 
Our benchmark offers a substrate where such existing and new methods and architectures can be integrated and systematically evaluated.

Following prior MU work \citep{kurmanji2023unboundedmachineunlearning, fan2024salunempoweringmachineunlearning,jia2023model}, 
we evaluate five representative methods spanning three common unlearning families: 
(i) fine-tuning-based methods, including Fine-tune and L1-Sparse, which update the model using retain data and are often effective for low-memorization examples; 
(ii) gradient-based methods, including NegGrad and NegGrad+, which explicitly promote forgetting through gradient ascent on $D_f$; 
and (iii) saliency-based methods, represented by SalUn, which performs parameter-selective unlearning based on saliency.
All MU methods assume a training dataset $D$ partitioned into a forget set $D_f$ and a retain set $D_r = D \setminus D_f$. 
Fine-tune continues training only on $D_r$, while L1-Sparse adds sparsity regularization to encourage sparse parameter updates.
NegGrad performs gradient ascent on $D_f$, whereas
NegGrad+ combines gradient ascent on $D_f$ with standard fine-tuning on $D_r$ to balance forgetting and retention simultaneously during unlearning. 
SalUn 
first identifies the key parameters influencing examples in $D_f$ using a saliency criterion, and then updates only these parameters while perturbing the associated labels of examples in $D_f$.

In addition to comparing raw instantiations of these methods, we instantiate all three within the RUM framework, which has been shown to consistently strengthen a wide range of MU algorithms \citep{zhao2024makesunlearninghard}. 
This allows us to compare method families under their best-performing configurations.

\subsection{Leveraging Memorization for Unlearning}
Memorization measures how strongly a trained model depends on a particular training example \citep{feldmanfirst, feldmanest}. 
Let $\mathcal{Z}=\mathcal{X}\times\mathcal{Y}$ denote the example space, and let $D=\{(x_j,y_j)\}_{j=1}^n \in \mathcal{Z}^n$ be the training set. 
Let $g_\theta$ be the classifier with parameters $\theta$ obtained by applying a training algorithm $\mathcal{A}$ on a training dataset $D$.
For an example $(x_i,y_i)\in D$,
let $D_{-i} := D \setminus {(x_i, y_i)}$ and $\Pr_D[\cdot] := \Pr_{f \sim A(D)}[\cdot]$, 
the measure, referred to as \textit{memorization score}, is defined as
\begin{equation}
\mathrm{mem}(A, D, i)
= \Pr_D[f(x_i)=y_i] - \Pr_{D_{-i}}[f(x_i)=y_i],
\end{equation}
capturing how predictions change when removing an example from training. Robust computation of this metric necessitates training a large number of models (each retrained on subsets excluding specific examples) to achieve stable estimates, which is especially expensive, particularly for VTs.
Proxies like the four \textit{Learning Events Proxies} (Confidence (Conf), Max Confidence (MaxConf), Entropy (Ent), and Binary Accuracy (BA)) \citep{jiang2021characterizingstructuralregularitieslabeled} and \textit{Holdout Retraining (HR)} \citep{carlini2019distributiondensitytailsoutliers} come to the rescue, efficiently estimating memorization scores. 
The proxy definitions and their correlations with Feldman memorization scores are given in Section \ref{app:mem}.

RUM leverages memorization to improve unlearning performance. It is structured in three stages:
(1) \textit{Refinement}: Partitions the forget set into homogeneous subsets based on memorization scores of examples in the forget set. Specifically, it creates three partitions consisting of low-, medium- and high-memorization forget examples.
(2) \textit{Matching}: Selects suitable unlearning methods tailored to each partition.  
(3) \textit{Unlearning}: Applies an unlearning algorithm sequentially on the three partitions.

We evaluate Fine-tune, NegGrad+, and SalUn within the RUM framework. Concretely, RUM applies each base unlearning method sequentially to the three memorization-based partitions (e.g., RUM(SalUn)). 
For simplicity, unless explicitly stated as ``vanilla'', all results in Sections 4.3--4.6 use the RUM-integrated
version of the corresponding base method.
We adopt this evaluation protocol because instantiating MU algorithms within RUM has been shown to substantially strengthen their performance \citep{zhao2024makesunlearninghard}. 
For completeness, we additionally demonstrate in Section \ref{app:rum-vs-vanilla} that this improvement over vanilla instantiations also holds for Vision Transformers.

\section{Experimental Setup}

\subsection{Datasets and Architectures}
\label{subsec:implementation}
For each dataset, as described in Section \ref{sec:mu}, we split the training data $D$ into a forget set $D_f$ (the examples whose influence should be removed from the model) and a retain set $D_r=D\setminus D_f$ (the remaining training examples). 
The original model $\theta_o$ is trained or fine-tuned on the full
training set $D$, while the retrained reference model $\theta_r$ is trained using the same initialization and training protocol, but only on $D_r$. 
The unlearned model $\theta_u$ is obtained by applying an unlearning algorithm to $\theta_o$ using $D_f$ and $D_r$.
Thus, $\theta_r$ represents the ideal target of exact unlearning, i.e., the model that would have been obtained had the forget examples never been used.
We evaluate unlearning methods on four different image classification benchmarks that vary in dataset size, semantic complexity, and number of classes: CIFAR-10, CIFAR-100 \citep{cifars}, SVHN \citep{svhn}, and ImageNet-1K \citep{deng2009imagenet}, which are standard datasets commonly adopted in prior machine unlearning studies. 
CIFAR-10 and CIFAR-100 provide controlled settings with increasing class granularity, while SVHN is larger but semantically simpler.
To assess scalability to larger and more complex settings, we additionally benchmark on ImageNet-1K using its validation set.

For Vision Transformer architectures, we focus on ViT, Swin-T, and DINOv2 variants. Our primary benchmarks are ViT-Small, Swin-Tiny, and DINOv2-Small, which have broadly comparable parameter scales to ResNet-50 but differ in architectural design and training paradigm: ViT uses global self-attention, Swin-T introduces hierarchical structure and locality through shifted-window attention, and DINOv2 is a self-supervised Vision Transformer backbone. 
For CNN baselines, we follow the training protocols used in prior CNN-based unlearning benchmarks. 

To study the effect of model capacity, we additionally evaluate smaller and larger variants, including ViT-Tiny, ViT-Base, Swin-Small, and Swin-Base. This allows us to examine whether unlearning behavior varies systematically with model scale and architectural bias.

\subsection{Evaluation Metrics}
\label{subsec:evaluation-metrics}
\textbf{Unlearning Metrics.} 
For each experiment, we compare three models:
(i) the original model $\theta_o$,
(ii) the retrained model $\theta_r$, and
(iii) the unlearned model $\theta_u = U(\theta_o,D_f,D_r)$, obtained by applying an unlearning algorithm $U$.
The goal of unlearning is for $\theta_u$ to behave like $\theta_r$: 
it should remove the influence of $D_f$ to achieve high ``forgetting quality''(i.e., match $\theta_r$ on $D_f$) while preserving performance on $D_r$ and unseen test data $D_{\mathrm{test}}$.
We adopt two primary metrics from \citet{zhao2024makesunlearninghard} to assess this balance.
$\text{ToW}(\theta_u, \theta_r, D_f, D_r, D_{test})$ (ToW for short):
\begin{equation}
\begin{aligned}
\text{ToW} &=
(1 - \Delta a(\theta_u, \theta_r, D_f))
\cdot (1 - \Delta a(\theta_u, \theta_r, D_r)) \cdot (1 - \Delta a(\theta_u, \theta_r, D_{\text{test}}))
\end{aligned}
\end{equation}
where 
$a(\theta, D) = \frac{1}{|D|} \sum_{(x,y) \in D} \mathds{1}[f(x; \theta) = y]$ is the accuracy of a model $f$ parametrized by $\theta$ on $D$,
and $\Delta a(\theta_u, \theta_r, D) = |a(\theta_u, D) - a(\theta_r, D)|$ is the absolute difference in accuracy between  $\theta_u$ and the retrained-from-scratch model $\theta_r$ on $D$.
A higher ToW score means that $\theta_u$ more closely matches the retrained model across forget, retain, and test data.
The second metric, $\text{ToW-MIA}(\theta_u, \theta_r, D_f, D_r, D_{test})$ (ToW-MIA):
\begin{equation}
\begin{aligned}
\text{ToW\text{-}MIA} &=
(1 - \Delta m(\theta_u, \theta_r, D_f))
\cdot (1 - \Delta a(\theta_u, \theta_r, D_r)) \cdot (1 - \Delta a(\theta_u, \theta_r, D_{\text{test}}))
\end{aligned}
\end{equation}
where $m(\theta, D) = \frac{TN_{D}}{|D|}$ and 
$\Delta m(\theta_u, \theta_r, D) = |m(\theta_u, D) - m(\theta_r, D)|$.
In ToW-MIA, $m(\theta, D)$ accounts for ``forget quality'' using Membership Inference Attack (MIA) performance. To calculate this, as in \citet{fan2024salunempoweringmachineunlearning, zhao2024makesunlearninghard, jia2023model}, we train a binary classifier $C$ that classifies examples as ``in-training'' or ``out-of-training''. 
We simplify our notation $l(f(x; \theta), y)$ (the cross-entropy loss of a model with weights $\theta$ on example $x$ with label $y$) to $l(x, y)$. 
$C$ is trained on loss values from a balanced dataset $D_{\text{t}}^b = \{(l(x_i, y_i), y_i^b)\}$, where examples $x_i$ are drawn equally from the retain set $D_r$ (labelled as $y_i^b=1$ for ``training'') and the test set $D_{\text{test}}$ (labelled as $y_i^b=0$ for ``non-training''). Once trained, $C$ evaluates loss values for examples in $D_f$. 
$m(\theta, D_f)$ measures the proportion of $D_f$ examples that $C$ classifies as ``non-training'', i.e. the true negatives $TN_{D_f}$ on $D_f$.
The hope is for the unlearning method to cause the model to treat forget set examples as if they were never seen during training, resulting in the classifier categorising them as ``non-training.'' Thus, similarly to $\Delta a$,  $\Delta m(\theta_u, \theta_r, D)$ quantifies how closely the unlearned model $\theta_u$ resembles the MIA performance of the retrain-from-scratch model $\theta_r$.

ToW and ToW-MIA measure ``forget quality'' differently: ToW uses accuracy differences on the forget set, while ToW-MIA uses differences in MIA vulnerability. Both range from 0 to 1, with higher values indicating better unlearning performance (a closer match to retraining from scratch).
Together, ToW and ToW-MIA provide a comprehensive view of unlearning performance.

\textbf{Memorization Proxy Metric.} 
\label{subsec:proxy-evaluation}
To evaluate how well a proxy captures memorization, we compute Spearman’s rank correlation coefficient between the ground-truth memorization score and the proxy value across training examples.
For each example $(x_i, y_i) \in D$, let $m_i = \text{mem}(A, D, i)$ denote the ``Feldman'' memorization score \citep{feldmanfirst} and $p_i = \text{proxy}(x_i, y_i)$ the value of a given proxy metric. We rank ${m_i}$ and ${p_i}$ independently and compute Spearman’s $\rho$ between their ranks.
The resulting coefficient $\rho \in [-1, 1]$ measures the strength and direction of the monotonic relationship between true memorization and the proxy, with larger absolute values indicating stronger correlation.



\section{Results and Analyses}
\label{sec:exp_ana}
In this section we present the main results, focusing on CIFAR-100 as a representative dataset.
The Appendix contains results for the other datasets.
Also, as we focus on memorization-based MU algorithms, we first establish that (i) the memorization patterns in VTs mimic those in CNNs and that (ii) the proxies are sufficiently reliable to be used by memorization-based MU algorithms in VTs.
The detailed results can be found in Appendix \ref{app:mem}.

\subsection{Memorization and Proxies in VTs}
Memorization plays a central role in unlearning:
a small subset of highly memorized examples may depend more strongly on their presence in the training set. 
We therefore first ask whether VTs memorize similarly to CNNs, and whether CNN-derived proxies remain valid predictors of memorization in VTs.  

\textbf{Key Takeaways--Memorization patterns.} 
We observe that VTs exhibit the same skewed, long-tailed memorization distributions previously reported for CNNs \citep{feldmanest}. As shown in Figure \ref{fig:memorization_hist_both} in Appendix \ref{app:mem_dist}, this pattern appears on both CIFAR-100 and CIFAR-10. 
On CIFAR-100, ViT-Small and Swin-Tiny closely resemble ResNet-50, suggesting that VTs and CNNs exhibit fundamentally similar memorization behavior despite architectural differences. 
On the simpler CIFAR-10 dataset, VTs show slightly lower mean memorization than ResNet-18, reflecting their ability (via pretraining and global attention) to rely less on memorizing individual examples for simpler tasks.
Overall, these results support the use of memorization-based unlearning for VTs.

\textbf{Key Takeaways--Proxy validity.} 
We next examine whether CNN-derived memorization proxies remain valid for VTs despite their different inductive biases and training regimes.
We evaluated five memorization proxies: Confidence (Conf), Max Confidence (MaxConf), Entropy (Ent), Binary Accuracy (BA) \citep{jiang2021characterizingstructuralregularitieslabeled}, and Holdout Retraining (HR) \citep{carlini2019distributiondensitytailsoutliers}; 
their detailed definitions are provided in Appendix \ref{subsec:proxy_corr}. 
As shown in Table \ref{tab:spearman_correlations},
Confidence achieves the strongest correlations across all models and datasets, with magnitudes (-0.79 to -0.91) close to those observed in CNNs.
Swin-Tiny shows slightly stronger correlations than ViT-Small, likely due to its hierarchical structure that more closely resembles traditional CNNs. 
HR shows moderate but significant positive correlations and is attractive in practice due to its large computational advantages vis-à-vis the other proxies. 
Overall, these results suggest that simple memorization proxies remain effective for VTs, enabling scalable memorization-based unlearning without expensive Feldman-score computation.

\begin{table}[h]
\centering
\caption{
Spearman correlations between memorization scores and proxies across CNNs and VTs. 
CNN-derived memorization proxies remain informative for VTs and can support scalable memorization-based unlearning.
}
\resizebox{0.72\columnwidth}{!}{%
    \begin{tabular}{lccccccc}
        \toprule
        & \multicolumn{3}{c}{\textbf{CIFAR-10}} & & \multicolumn{3}{c}{\textbf{CIFAR-100}} \\
        \cmidrule{2-4} \cmidrule{6-8}
        \textbf{Proxy} & \textbf{ResNet-18} & \textbf{ViT-Small} & \textbf{Swin-Tiny} & & \textbf{ResNet-50} & \textbf{ViT-Small} & \textbf{Swin-Tiny} \\
        \midrule
        \textbf{Conf}          & -0.80 & -0.79 & -0.88 & & -0.91 & -0.85 & -0.90 \\
        MaxConf      & -0.76 & -0.77 & -0.85 & & -0.87 & -0.80 & -0.86 \\
        Ent             & -0.75 & -0.78 & -0.85 & & -0.80 & -0.77 & -0.82 \\
        BA     & -0.71 & -0.63 & -0.79 & & -0.89 & -0.69 & -0.78 \\
        \textbf{HR}  & +0.67  & +0.45  & +0.64  & & +0.62  & +0.50  & +0.52  \\
        \bottomrule
    \end{tabular}
}
\label{tab:spearman_correlations}
\end{table}

\subsection{Extending Memorization-Based Unlearning to Vision Transformers}\label{app:rum-vs-vanilla}
Prior work has shown that incorporating memorization signals (e.g. RUM framework proposed by \citet{zhao2024makesunlearninghard}) can significantly improve the performance of a wide range of unlearning algorithms, particularly in CNN-based models \citep{zhao2024makesunlearninghard}. A natural question is whether these gains transfer to Vision Transformers, and whether memorization-based frameworks can similarly enhance existing methods in this setting.
To investigate this, we conduct an ablation study comparing vanilla unlearning algorithms against their RUM-integrated counterparts. Specifically, we evaluate Fine-tune, NegGrad+, and SalUn, along with a VT-specific method, LetheViT \citep{tong2025lethevitselectivemachineunlearning}, each with and without RUM. Experiments are performed on CIFAR-100 using ViT-Small and Swin-Tiny architectures.

Table \ref{tab:rum-vanilla} provides a controlled comparison for assessing the benefit of incorporating memorization signals. 
Across all methods and architectures, the RUM-based variants consistently outperform (or remain competitive with) their vanilla counterparts in both ToW and ToW-MIA, indicating more effective forgetting with improved resistance to membership inference.
This pattern holds for both ViT-Small and Swin-Tiny, showing that the gains from RUM are robust across different VT architectures.


\textbf{Key Takeaways.} 
Memorization-based unlearning (RUM) consistently improves existing algorithms in Vision Transformers, indicating that its benefits generalize beyond CNNs and provide a reliable mechanism for enhancing unlearning performance across architectures.
\begin{table}[t]
\centering
\caption{
Comparison of RUM-integrated and vanilla unlearning methods on CIFAR-100 using the Holdout Retraining proxy. Each entry reports \textbf{ToW ($\uparrow$) / ToW-MIA ($\uparrow$)} as mean $\pm$ 95\% confidence interval, with best results for each entry shown in bold. Across all architectures, RUM generally improves the base unlearning algorithms.
}
\label{tab:rum-vanilla}
\scriptsize
\setlength{\tabcolsep}{2.5pt}
\resizebox{\columnwidth}{!}{%
\begin{tabular}{lcccccc}
\toprule
& \multicolumn{2}{c}{\textbf{ViT-Small}} 
& \multicolumn{2}{c}{\textbf{Swin-Tiny}} 
& \multicolumn{2}{c}{\textbf{DINOv2-Small}} \\
\cmidrule(lr){2-3} \cmidrule(lr){4-5} \cmidrule(lr){6-7}
Method 
& \makecell{Within RUM \\ \scriptsize ToW ($\uparrow$) / ToW-MIA ($\uparrow$)}
& \makecell{Vanilla \\ \scriptsize ToW ($\uparrow$) / ToW-MIA ($\uparrow$)}
& \makecell{Within RUM \\ \scriptsize ToW ($\uparrow$) / ToW-MIA ($\uparrow$)}
& \makecell{Vanilla \\ \scriptsize ToW ($\uparrow$) / ToW-MIA ($\uparrow$)}
& \makecell{Within RUM \\ \scriptsize ToW ($\uparrow$) / ToW-MIA ($\uparrow$)}
& \makecell{Vanilla \\ \scriptsize ToW ($\uparrow$) / ToW-MIA ($\uparrow$)} \\
\midrule
Original 
& $0.619{\ci{0.040}}$ / $0.538{\ci{0.023}}$ & Same as left 
& $0.670{\ci{0.027}}$ / $0.586{\ci{0.025}}$ & Same as left
& $0.555{\ci{0.112}}$ / $0.425{\ci{0.080}}$ & Same as left \\

Fine-tune 
& $\mathbf{0.855}{\ci{0.036}}$ / $\mathbf{0.889}{\ci{0.008}}$ 
& $0.807{\ci{0.025}}$ / $0.844{\ci{0.023}}$
& $\mathbf{0.822}{\ci{0.034}}$ / $\mathbf{0.839}{\ci{0.032}}$ 
& $0.774{\ci{0.006}}$ / $0.781{\ci{0.020}}$
& $\mathbf{0.846}{\ci{0.038}}$ / $\mathbf{0.824}{\ci{0.070}}$ 
& $0.796{\ci{0.061}}$ / $0.768{\ci{0.064}}$ \\

NegGrad+ 
& $\mathbf{0.931}{\ci{0.039}}$ / $0.838{\ci{0.024}}$ 
& $0.888{\ci{0.022}}$ / $\mathbf{0.845}{\ci{0.026}}$
& $\mathbf{0.975}{\ci{0.027}}$ / $\mathbf{0.902}{\ci{0.035}}$ 
& $0.851{\ci{0.018}}$ / $0.831{\ci{0.012}}$
& $\mathbf{0.937}{\ci{0.044}}$ / $\mathbf{0.896}{\ci{0.032}}$ 
& $0.827{\ci{0.064}}$ / $0.762{\ci{0.044}}$ \\

SalUn 
& $\mathbf{0.903}{\ci{0.049}}$ / $\mathbf{0.709}{\ci{0.085}}$ 
& $0.884{\ci{0.046}}$ / $0.686{\ci{0.047}}$
& $\mathbf{0.870}{\ci{0.039}}$ / $\mathbf{0.751}{\ci{0.043}}$ 
& $0.833{\ci{0.093}}$ / $0.640{\ci{0.109}}$
& $\mathbf{0.831}{\ci{0.121}}$ / $\mathbf{0.840}{\ci{0.125}}$ 
& $0.736{\ci{0.146}}$ / $0.662{\ci{0.155}}$ \\

LetheViT 
& $\mathbf{0.848}{\ci{0.007}}$ / $\mathbf{0.879}{\ci{0.028}}$ 
& $0.808{\ci{0.047}}$ / $0.839{\ci{0.045}}$
& $\mathbf{0.815}{\ci{0.018}}$ / $\mathbf{0.851}{\ci{0.032}}$ 
& $0.773{\ci{0.014}}$ / $0.780{\ci{0.024}}$
& $\mathbf{0.784}{\ci{0.047}}$ / $\mathbf{0.743}{\ci{0.047}}$ 
& $0.727{\ci{0.030}}$ / $0.681{\ci{0.011}}$ \\
\bottomrule
\end{tabular}}
\end{table}

\subsection{Performance of Unlearning Algorithms on Vision Transformers}
\label{subsec:def_unlearning}
For these experiments, $D_f$ comprises 3,000 examples, divided into $M=3$ partitions of $N = 1,000$ examples each, representing the lowest, medium and highest proxy values. We apply each unlearning algorithm in the order of low → medium → high memorization, using memorization-proxy values. 
All results are averaged over three runs and we also present 95\% confidence intervals.

Given that pretrained transformer models have lower memorization, we anticipated that $\theta_r$ might already perform well on $D_f$. To establish a baseline for comparison, we calculated ToW and ToW-MIA parametrized by $(\theta_o, \theta_r, D_f, D_r, D_{test})$ denoted as ``Original''. 
This baseline indicates the performance we would achieve without applying any unlearning algorithm. 
Hyper-parameters for all algorithms are detailed in Appendix \ref{app:unlearn_hyper}.

\begin{figure*}[h]
  \centering
    \begin{subfigure}[b]{0.43\textwidth} 
        \centering
        \includegraphics[width=0.9\linewidth]{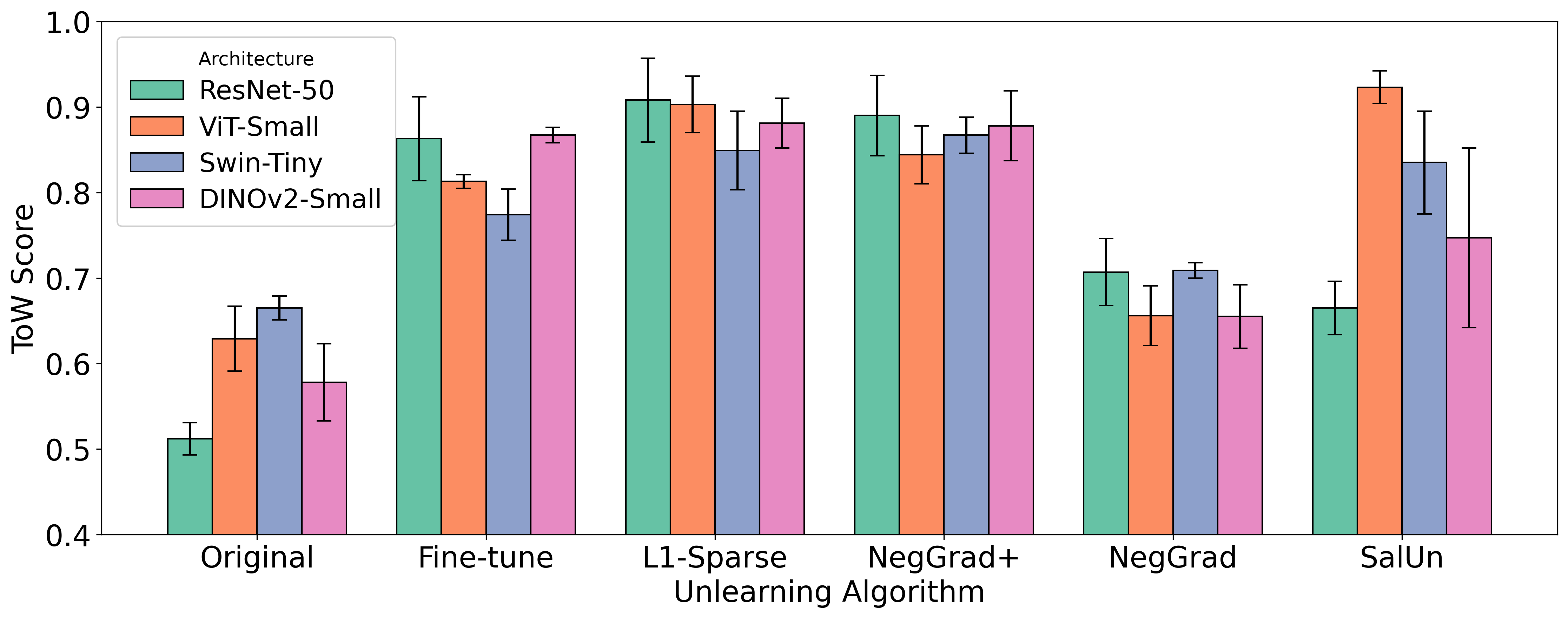}
        \caption{ToW with Confidence}
        \label{fig:main-fig-distr-verbatim}
    \end{subfigure}
    \begin{subfigure}[b]{0.43\textwidth}
        \centering
        \includegraphics[width=0.9\linewidth]{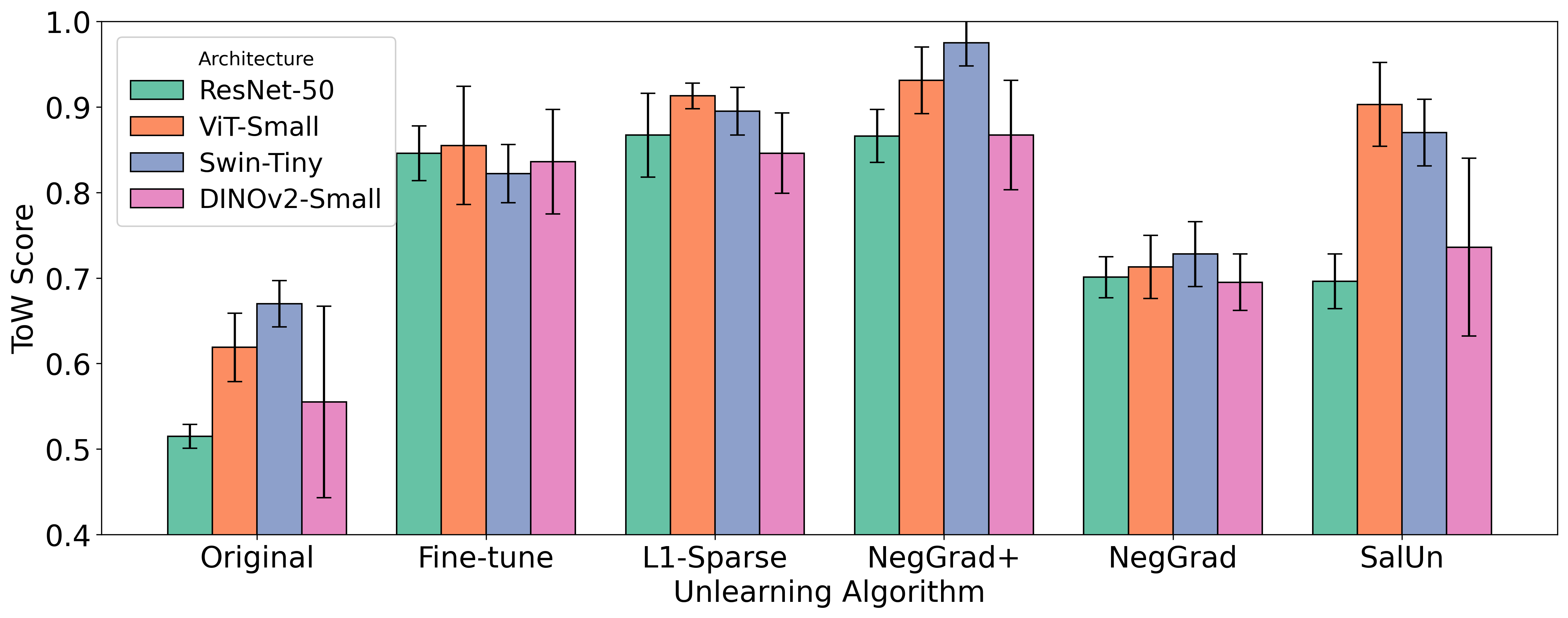}
        \caption{ToW with Holdout Retraining}
        \label{fig:main-fig-distr-template}
    \end{subfigure}

    \begin{subfigure}[b]{0.43\textwidth}
      \centering
      \includegraphics[width=0.9\linewidth]{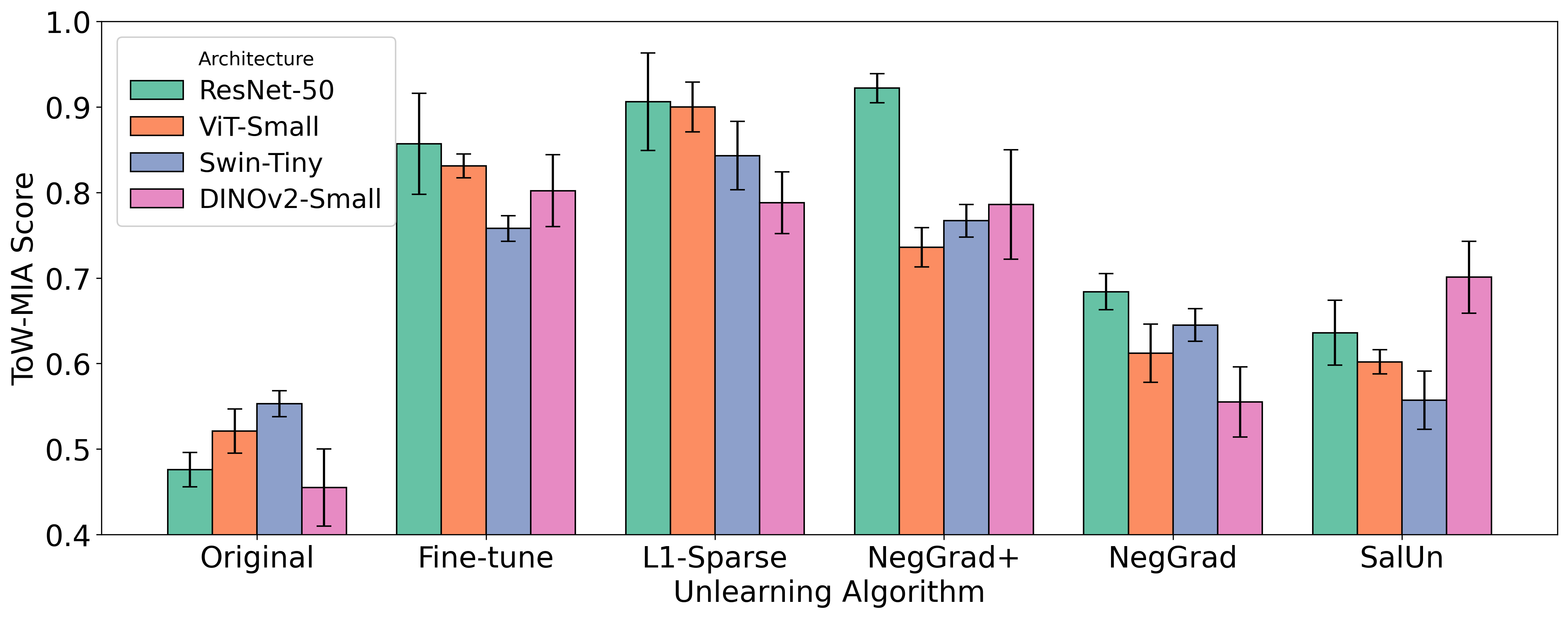}
      \subcaption{ToW-MIA with Confidence}
    \end{subfigure}
    \begin{subfigure}[b]{0.43\textwidth}
      \centering
      \includegraphics[width=0.9\linewidth]{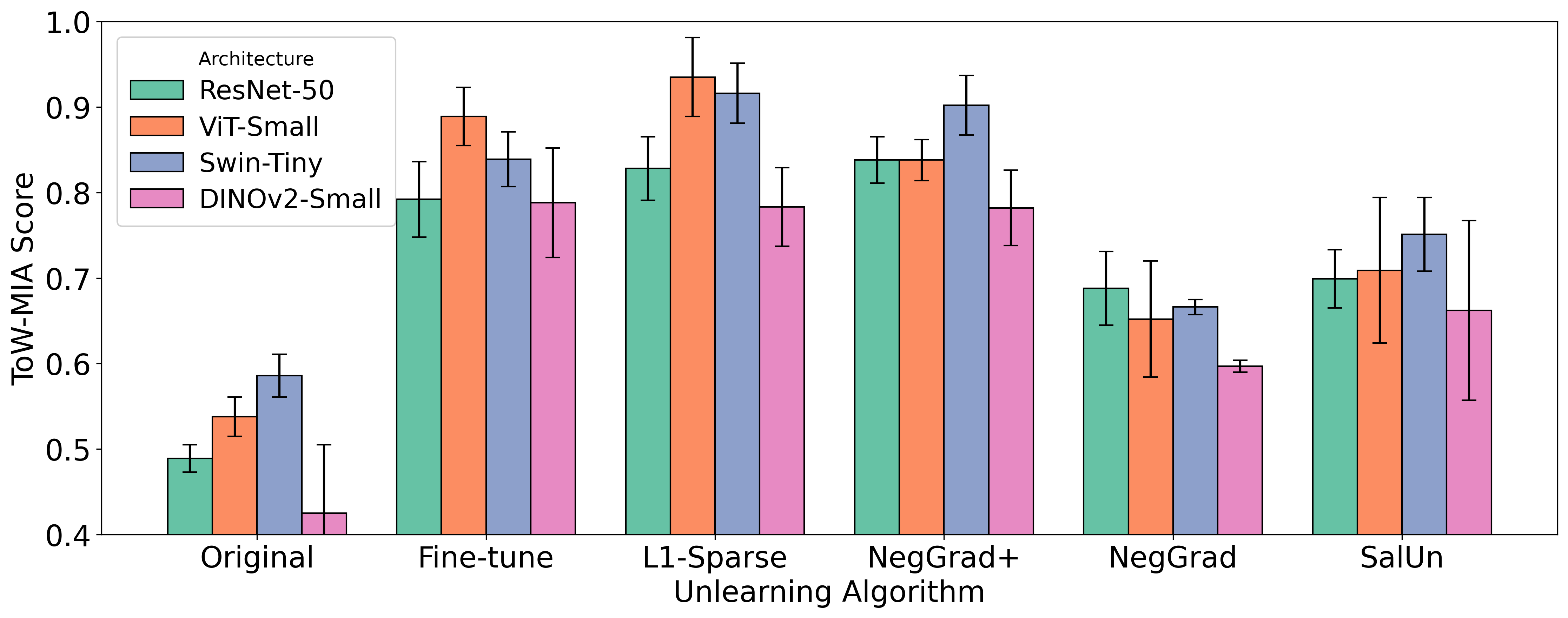}
      \subcaption{ToW-MIA with Holdout Retraining}
    \end{subfigure} \\
  \caption{
  Unlearning performance comparison on CIFAR-100, measured by ToW and ToW-MIA (higher is better), reported as mean with 95\% confidence interval.
  NegGrad+ is consistently strong, especially when paired with Holdout Retraining, while SalUn often achieves competitive ToW but weaker ToW-MIA.
  ViT favors fine-tuning-based methods (Fine-tune, L1-Sparse), whereas Swin-T favors gradient-based methods (NegGrad+, NegGrad).
  }
  \label{fig:cifar100_comparison}
\end{figure*}

\textbf{How do different MU approaches perform on VTs?}
Figure \ref{fig:cifar100_comparison} 
presents unlearning results across SOTA algorithms on CIFAR-100 (see Figures \ref{fig:cifar10_unlearning_comparison} and \ref{fig:svhn_unlearning_comparison} in Appendix for CIFAR-10 and SVHN).
Our results show clear family-level trends. Fine-tuning-based methods perform strongly on ViT, while gradient-based methods are more effective on Swin-T. Among them, 
NegGrad+ is consistently strong, especially on more complex datasets (CIFAR-100, and ImageNet-1K in Section \ref{subsec:imagenet_val_results}) and performs  best when paired with Holdout Retraining. Notably, NegGrad+ with Holdout Retraining on Swin-T even outperforms the corresponding ResNets baselines. 
SalUn achieves good ToW scores but struggles with ToW-MIA on more complex datasets.
From the proxy viewpoint, Holdout Retraining performs excellently compared to Confidence for CIFAR datasets, while Confidence can match it for the simpler SVHN. More detailed results are in Table \ref{tab:all_unlearn_metrics}.

\textbf{Key Takeaways.} 
CNN-derived MU methods can be equally (if not more) effective for VTs. NegGrad+ and L1-Sparse are the most robust MU methods for VTs
, while SalUn is vulnerable in ToW-MIA on harder datasets. Proxy choice (Holdout Retraining vs. Confidence) further shapes outcomes.


\textbf{How does unlearning performance in VTs compare to CNNs?}
We first compare the accuracy of the retrained model $\theta_r$ on $D_f$ in ResNets versus VTs, since this directly influences ToW and ToW-MIA evaluation for the benchmark model $\theta_r$.
Table \ref{tab:retrain_architectures} shows that on CIFAR-10, $\theta_r$ achieves much higher $D_f$ accuracy in VTs than CNNs (e.g., $\sim90\%$ for ViT/Swin-T/DINOv2 vs. $\sim50\%$ for ResNet-18). 
This advantage stems from VT pretraining, which enables learning robust feature representations. 
Thus, after $D_f$ examples are removed and the model is retrained, $\theta_r$ remains close to the original $\theta_o$ for VTs, 
explaining their high baseline ``Original'' performance (Table \ref{tab:all_unlearn_metrics}). 
However, this pretraining advantage diminishes on the more complex datasets (CIFAR-100).

Figure \ref{fig:cifar100_comparison} 
(and Figure \ref{fig:cifar10_unlearning_comparison} in Appendix) shows how method rankings shift across architectures and proxies.
With the Confidence proxy, Fine-tune and NegGrad+ perform best in ResNets, while VTs lag slightly. Under Holdout Retraining, however, VTs close or surpass this gap, particularly for NegGrad+.
SalUn behaves differently: it achieves good ToW scores in VTs but underperforms on ToW-MIA compared to CNNs, suggesting that while it effectively adjusts transformer outputs to match $\theta_r$, it struggles to protect against MIAs.

\textbf{Key Takeaways.} Pretraining gives VTs an advantage on simpler tasks, but this weakens with increasing complexity. 
MU method rankings from CNNs do not transfer to VTs; they depend strongly on architectures and proxies. 
Holdout Retraining narrows or reverses the gap in favor of VTs.
\begin{table}[htb]
  \centering
  \caption{
  $D_f$ accuracies of the retrained reference model $\theta_r$ across architectures. 
  VTs show much higher $D_f$ accuracy than ResNet-18 on CIFAR-10, 
  but this gap largely disappears on the more complex CIFAR-100 setting.
}
  \resizebox{0.7\linewidth}{!}{%
  \begin{tabular}{@{}lcc|cc@{}}
    \toprule
    \multirow{2}{*}{\textbf{Architecture}} 
    & \multicolumn{2}{c|}{\textbf{CIFAR-10}} 
    & \multicolumn{2}{c}{\textbf{CIFAR-100}} \\
    & \textbf{Confidence} & \textbf{Holdout Retraining}
    & \textbf{Confidence} & \textbf{Holdout Retraining} \\
    \midrule
    ResNet-18  & $50.433 \ci{6.808}$ & $62.922 \ci{4.681}$ & -- & -- \\
    ResNet-50  & -- & -- & $64.267 \ci{0.504}$ & $69.856 \ci{2.620}$ \\
    ViT-Small  & $94.089 \ci{0.456}$ & $89.244 \ci{1.321}$ & $69.322 \ci{1.788}$ & $68.767 \ci{3.828}$ \\
    Swin-Tiny  & $91.867 \ci{1.711}$ & $86.389 \ci{1.246}$ & $69.833 \ci{1.242}$ & $70.267 \ci{2.848}$ \\
    DINOv2-Small  & $88.529 \ci{2.101}$ & $92.333 \ci{3.234}$ & $58.267 \ci{3.770}$ & $58.778 \ci{4.944}$ \\
    \bottomrule
  \end{tabular}
  }
  \label{tab:retrain_architectures}
\end{table}

\subsection{How do VT architectures and capacities affect MU performance?}
\label{subsec:diff_trans}

We observe systematic differences among ViT, Swin-T, and DINOv2, as well as clear effects of model size. 

\textbf{Architecture.} 
Swin-T exhibits stronger memorization than ViT (e.g., higher mean Feldman scores, heavier tails on CIFAR-10: $\mu $=0.17 vs. 0.09, see Figure 
\ref{fig:memorization_hist_both}), which likely explains its superior performance with gradient-based methods such as NegGrad+. 
Swin-T also aligns with CNN-like behavior in SalUn, achieving optimal performance at saliency threshold $\gamma=0.3$, consistent with CNNs \citep{zhao2024makesunlearninghard}, whereas ViT requires a much lower $\gamma=0.1$. 
Together, these results suggest that ViT’s global attention leads to more diffuse parameter involvement, while Swin-T’s local windowed attention allows for more concentrated, targeted unlearning.
Representation-level analyses in Appendix \ref{app:representation_analysis} further support this view: embedding visualizations show distinct forget/retain shifts after unlearning, and CKA shows that changes concentrate in later representations, with sharper late-layer sensitivity in ViT and a more staged pattern in Swin-T.
DINOv2, as a self-supervised backbone with global attention, generally follows trends closer to ViT than Swin-T, while showing less sensitivity to the choice of unlearning algorithm, maintaining competitive results across Fine-tune, NegGrad+, and SalUn.

In terms of algorithm performance, Figure \ref{fig:cifar100_comparison} (and Figure \ref{fig:cifar10_unlearning_comparison}, Table \ref{tab:all_unlearn_metrics} 
in Appendix) shows that 
fine-tuning-based methods are
particularly effective on ViT
while gradient-based methods excel
on Swin-T, especially with Holdout Retraining on more complex tasks.
DINOv2 achieves competitive performance across algorithms, though typically without matching the peak performance observed for Swin-T under gradient-based methods or ViT under fine-tuning.
Saliency-based unlearning
consistently attains good ToW but struggles on ToW-MIA, especially for ViT-Small (e.g., 0.582 on CIFAR-10 with Confidence).

Dataset-level effects mirror these trends: ViT has an edge on the smaller, medium-complexity CIFAR-10, Swin-T dominates on the larger, more complex CIFAR-100, 
and DINOv2 remains consistently competitive across datasets.
All architectures perform well on the simpler SVHN dataset.
Continual unlearning results in Section \ref{subsec:stab_prox} show broadly similar trends across ViT and Swin-T. 

\textbf{Key Takeaways.} 
Architecture shapes which unlearning family works best: 
ViT favors fine-tuning-based methods, likely due to its global attention; Swin-T favors gradient-based unlearning 
likely due to its local, hierarchical attention. 
DINOv2 shows more balanced behavior across methods.
Dataset complexity further modulates these trends.

\textbf{Capacity.} 
To study the effect of model capacity, we extend our analysis to smaller and larger variants within each family, evaluated on CIFAR-10:
ViT-Tiny ($\sim 5.5$M params) vs.\ ViT-Small ($\sim 21.6$M) and Swin-Tiny ($\sim 27.5$M) vs.\ Swin-Small ($\sim 48.8$M). We focus on the Holdout Retraining (HR) proxy, which was especially promising in earlier results. We report ToW and ToW-MIA for the three MU algorithms (Fine-tune, NegGrad+, SalUn) alongside the Original baseline in Table \ref{tab:cifar10_holdout_4arch}.
The headline findings continue to hold: Fine-tune and NegGrad+ remain consistently strong; SalUn attains high ToW but is often much weaker on ToW-MIA and is notably sensitive to model size, architecture, and proxy choice.
We also observe \textit{architecture-specific capacity trends:} 
For Swin-T, increasing capacity from Tiny to Small offers little ToW improvement and can reduce ToW-MIA, suggesting Swin-Tiny is already sufficient for CIFAR-10 and that Swin-Small may overfit. For ViT, the opposite holds: ViT-Tiny underperforms ViT-Small on both ToW and ToW-MIA, implying under-capacity limits MU effectiveness. 
The results reveal a ``sweet spot'' around ViT-Small and Swin-Tiny, where models are neither under- nor overfitting, yielding 
a better balance between unlearning efficacy and privacy.

\begin{table}[h]
\centering
\caption{
  Effect of VT capacity on CIFAR-10 using the HR proxy. Results are reported as mean $\pm$ 95\% confidence interval for ToW ($\uparrow$) and ToW-MIA ($\uparrow$). ViT-Small and Swin-Tiny provide the best overall balance, suggesting a capacity ``sweet spot'' for unlearning.
}
\label{tab:cifar10_holdout_4arch}
\scriptsize
\setlength{\tabcolsep}{2.5pt}
\resizebox{0.85\columnwidth}{!}{%
\begin{tabular}{lcc@{\hspace{0.8em}}cc}
\toprule
Algorithm
& \makecell{ViT-Small\\[0.15em]\scriptsize ToW ($\uparrow$) / ToW-MIA ($\uparrow$)}
& \makecell{ViT-Tiny\\[0.15em]\scriptsize ToW ($\uparrow$) / ToW-MIA ($\uparrow$)}
& \makecell{Swin-Small\\[0.15em]\scriptsize ToW ($\uparrow$) / ToW-MIA ($\uparrow$)}
& \makecell{Swin-Tiny\\[0.15em]\scriptsize ToW ($\uparrow$) / ToW-MIA ($\uparrow$)} \\
\midrule
Original
& $\mathbf{0.891}{\ci{0.016}}$ / $\mathbf{0.831}{\ci{0.014}}$
& $0.773{\ci{0.020}}$ / $0.699{\ci{0.020}}$
& $\mathbf{0.886}{\ci{0.013}}$ / $\mathbf{0.832}{\ci{0.020}}$
& $0.862{\ci{0.008}}$ / $0.811{\ci{0.018}}$ \\

Fine-tune
& $\mathbf{0.928}{\ci{0.013}}$ / $\mathbf{0.913}{\ci{0.008}}$
& $0.862{\ci{0.060}}$ / $0.854{\ci{0.060}}$
& $0.921{\ci{0.005}}$ / $0.924{\ci{0.010}}$
& $\mathbf{0.923}{\ci{0.021}}$ / $\mathbf{0.931}{\ci{0.021}}$ \\

NegGrad+
& $0.916{\ci{0.016}}$ / $\mathbf{0.968}{\ci{0.015}}$
& $\mathbf{0.957}{\ci{0.011}}$ / $0.869{\ci{0.011}}$
& $0.944{\ci{0.028}}$ / $0.888{\ci{0.021}}$
& $\mathbf{0.977}{\ci{0.023}}$ / $\mathbf{0.924}{\ci{0.028}}$ \\

SalUn
& $\mathbf{0.956}{\ci{0.003}}$ / $\mathbf{0.766}{\ci{0.039}}$
& $0.846{\ci{0.043}}$ / $0.640{\ci{0.043}}$
& $0.950{\ci{0.031}}$ / $0.568{\ci{0.093}}$
& $\mathbf{0.961}{\ci{0.048}}$ / $\mathbf{0.834}{\ci{0.022}}$ \\
\bottomrule
\end{tabular}}
\end{table}

\textbf{Key Takeaways.} Performance trends remain stable across ViT-Tiny/Small and Swin-Tiny/Small. 

\subsection{Results on Larger / More Complex Data}
\label{subsec:imagenet_val_results}

To examine unlearning under larger-scale and more complex conditions, we leverage that our Vision Transformers are pretrained on ImageNet-1K and use the \emph{validation} split (50{,}000 images unseen during pretraining) as a new evaluation dataset (already $5\times$ larger than CIFAR-100’s test set). We form a forget set of $|D_f|{=}3{,}000$ (partitioned into three 1k subsets for low/medium/high proxy values, as in Section \ref{subsec:def_unlearning}), and treat the remaining 47k images as retain/test set. As this setup has only retain/test and forget components, we report ToW with two terms (retain/test accuracy and forget accuracy). 
We evaluate on 
Swin-Small ($\sim$49M params), as well as larger VT variants, Swin-Base ($\sim$88M parameters) and ViT-Base/16 ($\sim$85M parameters), and use the Holdout Retraining proxy since it was previously found to be high performing.
\begin{table}[htb]
\centering
\caption{
Unlearning performance measured by ToW ($\uparrow$) on ImageNet-1K validation set with HR proxy, comparing larger VT variants.
The method-level trends observed on CIFAR-100 persist in larger and more complex settings.
}
\label{tab:imagenet_val_model_scaling}
\footnotesize
\resizebox{0.6\linewidth}{!}{
\begin{tabular}{lccc}
\toprule
Method 
& Swin-Small (49M) 
& Swin-Base (88M) 
& ViT-Base/16 (85M) \\
\midrule
Fine-tune
& $0.780{\ci{0.023}}$
& $0.808{\ci{0.007}}$
& $0.843{\ci{0.016}}$ \\
NegGrad+
& $0.819{\ci{0.018}}$
& $0.837{\ci{0.004}}$
& $0.842{\ci{0.005}}$ \\
SalUn
& $0.743{\ci{0.027}}$
& $0.746{\ci{0.013}}$
& $0.773{\ci{0.027}}$ \\
LetheViT
& $0.733{\ci{0.017}}$
& $0.752{\ci{0.026}}$
& $0.773{\ci{0.030}}$ \\
\bottomrule
\end{tabular}
}
\end{table}

\begin{wraptable}{R}{5.5cm}
  \centering
  \caption{
  $D_f$ accuracies of the retrained model $\theta_r$ (with HR) on ImageNet-1K validation setting. Similar accuracy suggests that VT pretraining advantages diminish on more complex data.}
  \label{tab:imagenet_val_retrain_acc}
  \resizebox{0.8\linewidth}{!}{
  \begin{tabular}{lc}
    \toprule
    \textbf{Architecture} & \textbf{$D_f$ accuracy ($\%$)} \\
    \midrule
    ResNet-50   & $69.139 \ci{1.722}$ \\
    Swin-Small  & $69.433 \ci{2.246}$ \\
    \bottomrule
  \end{tabular}
  }
\end{wraptable}

Table \ref{tab:imagenet_val_model_scaling} shows that increasing model capacity within the Swin-T family leads to modest but consistent performance gains. Importantly, the qualitative trends observed at smaller scales persist: \emph{Fine-tune} and \emph{NegGrad+} remain strong performers, while \emph{SalUn} continues to underperform. 
This reinforces the conclusion that comparatively simple unlearning strategies can remain effective even for larger models and datasets, albeit performance varies with architecture and data complexity.

We also observe that absolute ToW values are lower here than those for Swin-Tiny on CIFAR-100, despite Swin-Small's larger capacity. This is reasonable: (i) retain/test accuracy is harder 
given ImageNet’s larger scale, diversity, and label space, and (ii) forgetting is more challenging on richer ImageNet images (greater embedding entanglement). Consequently, both terms 
in the two-factor ToW metric are lower than in the CIFAR-100 setting,
and further gains would likely require even larger-capacity models and/or stronger regularization.
Finally, Table \ref{tab:imagenet_val_retrain_acc} 
reinforces our earlier observation that the benefit of pretraining for forget accuracy diminishes as task complexity increases (cf. Table \ref{tab:retrain_architectures}).


\textbf{Key Takeaways.} 
Fine-tuning- and gradient-based baselines remain effective at ImageNet scale, whereas saliency-based unlearning underperforms, and the benefits of pretraining diminish as complexity grows.
Increasing model capacity at ImageNet scale yields small but consistent improvements, and 
method-family trends remain consistent.

\subsection{Does Continual Unlearning Impact Performance in Vision Transformers?}
\label{subsec:stab_prox}
In practical deployments, unlearning is often performed continually, through a sequence of smaller forget operations rather than a single large one. We therefore examine whether repeated unlearning leads to cumulative degradation in performance.
Based on earlier results, we focus on NegGrad+ with Holdout Retraining, which is the best-performing algorithm-proxy pair. 

We first perform five sequential unlearning steps, unlearning the same total number of examples ($|D_f| = 3{,}000$). At each step, proxy values are recomputed, and the forget set is partitioned as before, with $|D_f| = 600$ per step and $N = 200$ examples per partition. 
To further stress-test continual unlearning, we extend this to ten sequential steps under the same setting (NegGrad+ with HR), using both ViT-Small and Swin-Tiny.

\begin{figure}[htb]
  \centering
    \begin{subfigure}[b]{0.35\textwidth}
      \includegraphics[width=0.95\textwidth]{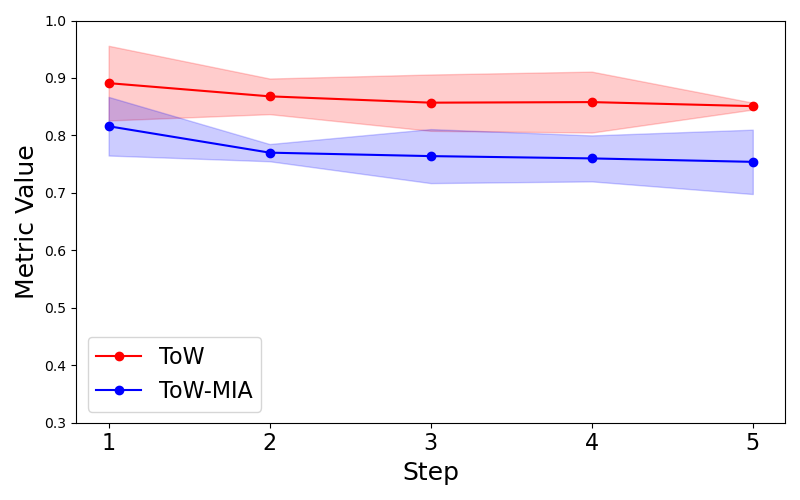}
      \subcaption{ViT-Small, 5 steps}
      \label{fig:lineplot_cifar100_vit_small}
    \end{subfigure}
    \begin{subfigure}[b]{0.45\textwidth}
      \includegraphics[width=0.9\textwidth]{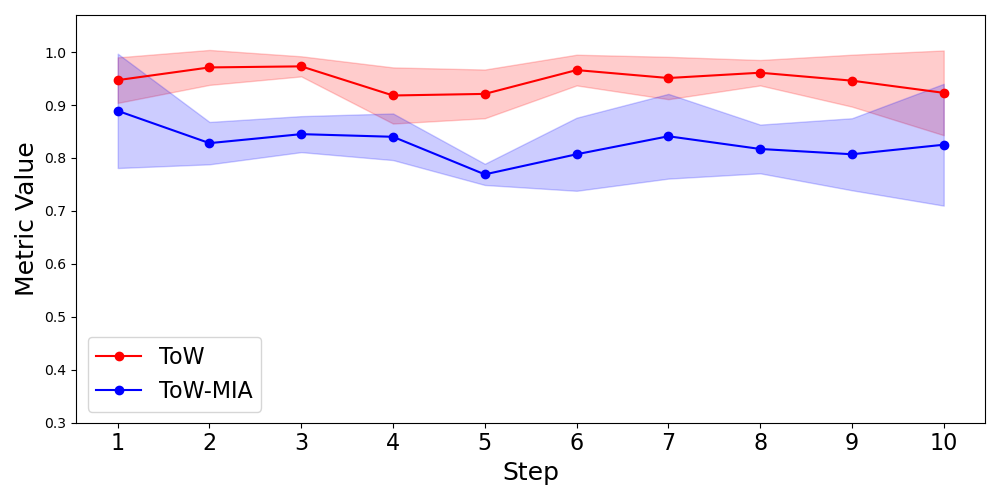}
      \subcaption{ViT-Small, 10 steps}
      \label{fig:lineplot_cifar100_swin_tiny}
    \end{subfigure}
  \caption{Continual unlearning performance across 5/10 Steps on CIFAR-100 using ViT-Small, reported with 95\% confidence intervals. 
  Performance remains stable across 5/10 steps; 
  results for Swin-Tiny and additional datasets are reported 
  in Figure \ref{fig:sequential_unlearning_10steps} 
  in Appendix \ref{app:continual_mu_cifar10}.
}
  \label{fig:sequential_unlearning_cifar100}
\end{figure}

As shown in Figure \ref{fig:sequential_unlearning_cifar100}, performance remains stable across steps, with no evidence of substantial accumulated degradation in either ToW or ToW-MIA. 
While ViT-Small exhibits a mild downward trend, all metrics remain within the confidence intervals of the initial unlearning step and are comparable to single-step results reported in Section \ref{subsec:def_unlearning}. 
Similar trends hold for Swin-Tiny, as well as for CIFAR-10 and SVHN (see Figure \ref{fig:sequential_unlearning_cifar10}, \ref{fig:sequential_unlearning_10steps}, Table \ref{tab:cu-10steps} in Appendix \ref{app:continual_mu_cifar10}).

\textbf{Key Takeaway.} Minimal (if any) degradation occurs under continual unlearning in VTs.

\section{Conclusions}
We benchmarked the performance of fundamentally different approaches to MU, ranging from simpler baselines to more advanced SOTA unlearning algorithms in VTs. We focused on algorithms leveraging memorization, as memorization has been found to be key to unlearning (across modalities) and the SOTA MU approaches for vision tasks can be substantially improved by leveraging it. 
To ensure the practicality of leveraging memorization, we studied memorization patterns in VTs and five different memorization proxies that have been found to offer high memorization-score fidelity in CNNs while being drastically more efficient to compute.
We employed different datasets to account for the effects of varying sizes and complexities. We also studied different popular VT architectures (ViT, Swin-T, and DINO) selected to 
span different degrees of similarity to CNN inductive biases.
We also used different model capacities to see the impact of model capacity on MU performance (with carefully selected dataset sizes to avoid over- and under-fitting).
This work represents the first comprehensive study in this domain and sheds new light on the current state of affairs of MU algorithm performance in VTs.

This work contributes the following novel insights: 
(i) Pretraining of VTs can help improve unlearning for smaller and simpler datasets/tasks. 
(ii) VTs and CNNs largely exhibit the same memorization patterns.
(iii) Well-known memorization proxies from CNNs can benefit unlearning performance in VTs as they can do in CNNs.
(iv) Continual unlearning does not degrade the efficacy of memorization proxies and overall unlearning performance in VTs.
(v) VTs can enjoy similar, if not better, unlearning performance to that in CNNs (using CNN-derived algorithms).
(vi) Swin-T can outperform ViT on more complex datasets, due to Swin-T's architectural similarities to CNNs.
(vii) NegGrad+, perhaps surprisingly, emerges as a consistent, strong performer for VTs, especially on more complex datasets.
(viii) Holdout Retraining emerges as the proxy yielding superior unlearning performance in VTs, whereas Confidence can be competitive on lower-complexity datasets.
We hope these insights, as well as the associated publicly available codebase/dataset infrastructure benchmark, shed ample light in a previously unstudied but important area and that 
they will serve as a foundation and comparison point for future studies of unlearning algorithms.






\bibliography{collas2026_conference}
\bibliographystyle{collas2026_conference}

\newpage 

\appendix
\section{Appendix}

\etocsettocdepth{subsubsection}
\localtableofcontents




\subsection{Implementation Details}
This section outlines the key implementation details for all experiments conducted in this study. 
For all experiments involving transformer architectures, we fine-tune models initialized with ImageNet-1K pretrained weights provided by the \texttt{timm} library \cite{rw2019timm}, as described in Section \ref{subsec:implementation}.

All experiments on the CIFAR-10 and CIFAR-100 datasets were conducted using a mix of NVIDIA RTX 2080 Ti, NVIDIA A10 and NVIDIA TITAN Xp GPUs, taking roughly 210 GPU hours. For the experiments on the SVHN dataset, training and evaluation were performed on NVIDIA A5000 GPUs, which consumed approximately 300 GPU hours.  

The code for reproducing the results is available at:
\url{https://github.com/kairanzhao/Unlearning_VTs}

\subsubsection{Memorization Estimation}
\label{app:memest}

For memorization estimation, we use ResNet-18 for CIFAR-10 and rely on precomputed memorization scores provided by  \cite{feldmanest} for CIFAR-100. Additionally, we compute memorization scores using ViT-Small and Swin-Tiny architectures for both CIFAR-10 and CIFAR-100. The training configurations and corresponding hyper-parameters are summarized in Table \ref{tab:train_config_est}.

\begin{table}[htb]
  \caption{Training configurations along with hyper-parameters of the models used to compute memorization scores shown in Figures 
  \ref{fig:memorization_hist_both}.
  The ``Transformers'' column represents the configuration for both ViT-Small and Swin-Tiny across CIFAR-10 and CIFAR-100.}
  \centering
  \begin{tabular}{lcc}
    \toprule
    & \textbf{ResNet-18} & \textbf{Transformers} \\
    \midrule
    \textbf{Optimizer}               & SGD                       & AdamW \\
    \textbf{Base learning rate}      & 0.01                       & 0.0001 \\
    \textbf{Loss}                    & Cross-Entropy             & Cross-Entropy \\
    \textbf{Learning rate scheduler} & Step decay                & CosineAnnealingLR \\
    \textbf{Batch size}              & 512                       & 128 \\
    \textbf{Epochs}                  & 30                        & 30 \\
    \textbf{Momentum}                & 0.9                       & - \\
    \textbf{Weight decay}            & $0.0005$                  & $0.05$ \\
    \textbf{Data augmentation}       & None                      & None \\
    \bottomrule
  \end{tabular}
  \label{tab:train_config_est}
\end{table}

\subsubsection{Unlearning}
This section presents the configuration and hyper-parameter details used to obtain the necessary models for ToW and ToW-MIA: ``Original'' $\theta_o$, ``Retrained'' $\theta_r$ and ``Unlearned'' $\theta_u$. 
\label{app:unlearn_hyper}
Table \ref{tab:train_config_train} shows the hyper-parameter details for training original models \& retraining. Table \ref{tab:unlearning_hyperparams} shows the unlearning hyper-parameters, and Table \ref{tab:unlearn_params} presents the sequential unlearning hyper-parameters.
\begin{table}[H]
  \caption{Training configurations and hyperparameter used for the original models ($\theta_o$) and the retrained models ($\theta_r$), using ViT-Small and Swin-Tiny architectures on CIFAR-10, CIFAR-100, and SVHN datasets.}
  \centering
  \begin{subtable}[t]{\columnwidth}
  \centering
  \begin{tabular}{lc}
    \toprule
    \textbf{Hyperparameter}       & \textbf{ViT-Small} \& \textbf{Swin-Tiny} \\
    \midrule
    \textbf{Optimizer}               & AdamW \\
    \textbf{Base learning rate}      & 0.0001 \\
    \textbf{Loss}                    & Cross-Entropy \\
    \textbf{Learning rate scheduler} & CosineAnnealingLR \\
    \textbf{Batch size}              & 128 \\
    \textbf{Epochs}                  & 50 \\
    \textbf{Weight decay}            & $0.05$ \\
    \textbf{Data augmentation}       &  Random Crop + Horizontal Flip \\
    \bottomrule
  \end{tabular}
  \caption{CIFAR-10, CIFAR-100}
  \end{subtable}
  \begin{subtable}[t]{\columnwidth}
  \centering
  \begin{tabular}{lcc}
    \toprule
    \textbf{Hyperparameter}       & \textbf{ViT-Small} & \textbf{Swin-Tiny} \\
    \midrule
    \textbf{Optimizer}               & AdamW & AdamW\\
    \textbf{Base learning rate}      & 0.0001 & 0.0001 \\
    \textbf{Loss}                    & Cross-Entropy & Cross-Entropy \\
    \textbf{Learning rate scheduler} & CosineAnnealingLR & CosineAnnealingLR\\
    \textbf{Batch size}              & 256 & 256 \\
    \textbf{Epochs}                  & 30 & 30 \\
    \textbf{Weight decay}            & 0.01 & 0.05\\
    \textbf{Data augmentation}       &  - &  - \\
    \bottomrule
  \end{tabular}
  \label{tab:svhn_config_train}
  \caption{SVHN}
\end{subtable}
  \label{tab:train_config_train}
\end{table}

\begin{table}[H]
  \caption{Hyper-parameters across all unlearn methods within the RUM$_F$ meta-algorithm on both Confidence and Holdout Retraining proxies to derive the unlearn models $\theta_u$. ``Unlearn Epochs'' column refers to the performed number of epochs each algorithm performed for the three memorization partitions (low $\rightarrow$ medium $\rightarrow$ high) respectively. }
  \centering
  \begin{tabular}{l l c c c c c}
    \toprule
    \textbf{Algorithm} & \textbf{Architecture} & \textbf{Unlearn Epochs} & \textbf{Unlearn LR} & \textbf{$\beta$} & \textbf{$\alpha$} & \textbf{$\gamma$} \\
    \midrule
    \multirow{2}{*}{Fine-tune} 
      & ViT-Small & 5,5,10 & 0.0001  & --  & --   & --  \\ 
      & Swin-Tiny & 5,5,10 & 0.0001  & --  & --   & --  \\ 
    \midrule
    \multirow{2}{*}{NegGrad+} 
      & ViT-Small & 5,5,10 & 0.00002 & 0.97  & --   & --  \\ 
      & Swin-Tiny & 5,5,10 & 0.00002 & 0.97  & --   & --  \\ 
    \midrule
    \multirow{2}{*}{SalUn} 
      & ViT-Small & 5,5,10 & 0.00005 & --  & 1 & 0.1 \\ 
      & Swin-Tiny & 5,5,10 & 0.0002  & --  & 1 & 0.3 \\ 
    \bottomrule
  \end{tabular}
  \label{tab:unlearning_hyperparams}
\end{table}


\begin{table}[H]
  \caption{Hyper-parameters at each sequential unlearning step for the NegGrad+ and Holdout Retraining configuration utilized. ``Unlearn Epochs'' are reduced to 1 for partition (low $\rightarrow$ medium $\rightarrow$ high) due to the smaller forget set size($|D_f|=600$) at each step. 
  All other settings remain consistent with the primary experiments.}
  \centering
  \begin{tabular}{ccc}
    \toprule
    \textbf{Unlearn Epochs} & \textbf{Unlearn LR} & \textbf{$\beta$} \\
    \midrule
    1,1,1                  & 0.00002             & 0.97            \\
    \bottomrule
  \end{tabular}

  \label{tab:unlearn_params}
\end{table}

\subsection{Memorization and Proxies Analysis in Vision Transformers}
\label{app:mem}

\subsubsection{Do Vision Transformers Memorize the Same Way as CNNs?}
\label{app:mem_dist}
As memorization plays a key role in unlearning, 
we first examine whether VTs exhibit memorization patterns similar to those observed in CNNs. 
For ResNet-18 on CIFAR-10 and both VT architectures on CIFAR-10/CIFAR-100, we compute memorization scores using the training configuration in Table \ref{tab:train_config_est}. For ResNet-50 on CIFAR-100, we use the precomputed scores from \citet{feldmanest}.

\begin{figure*}[htb]
  \centering

  \begin{subfigure}[t]{0.32\textwidth}
    \centering
    \includegraphics[width=\textwidth]{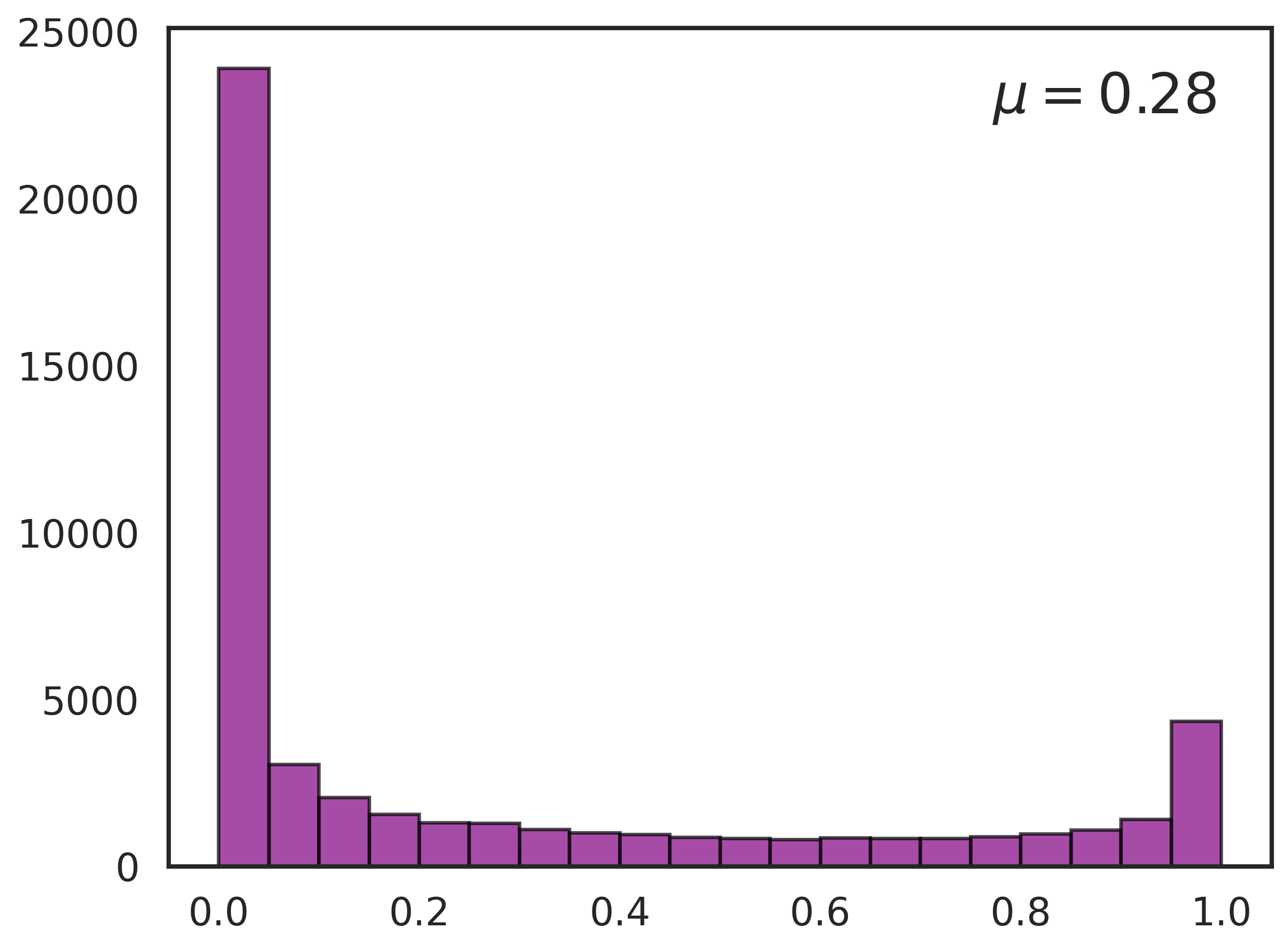}
    \subcaption{CIFAR-100, ResNet-50}
    \label{fig:memhist_c100_resnet}
  \end{subfigure}\hfill
  \begin{subfigure}[t]{0.32\textwidth}
    \centering
    \includegraphics[width=\textwidth]{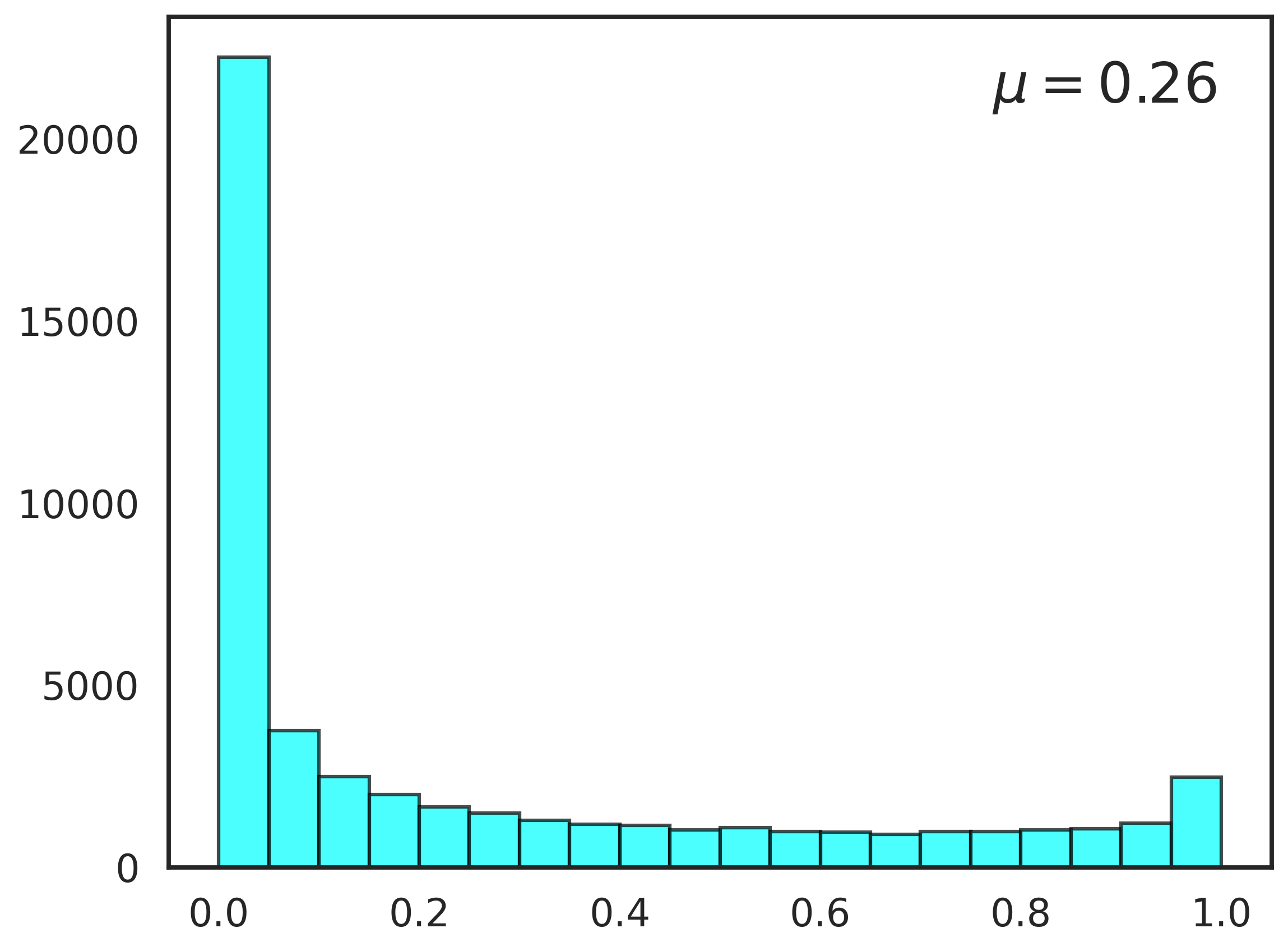}
    \subcaption{CIFAR-100, ViT-Small}
    \label{fig:memhist_c100_vit}
  \end{subfigure}\hfill
  \begin{subfigure}[t]{0.32\textwidth}
    \centering
    \includegraphics[width=\textwidth]{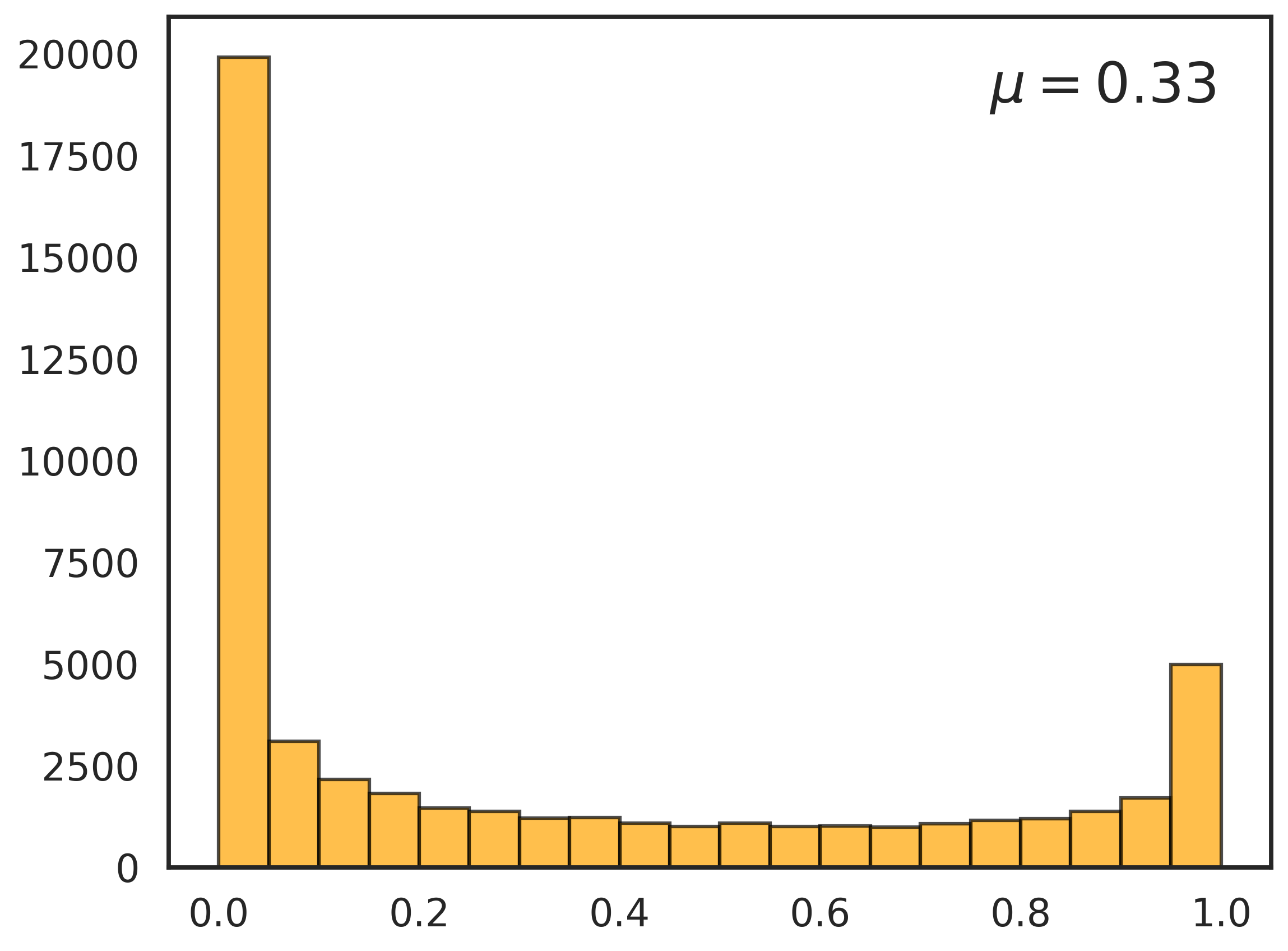}
    \subcaption{CIFAR-100, Swin-Tiny}
    \label{fig:memhist_c100_swin}
  \end{subfigure}


  \begin{subfigure}[t]{0.32\textwidth}
    \centering
    \includegraphics[width=\textwidth]{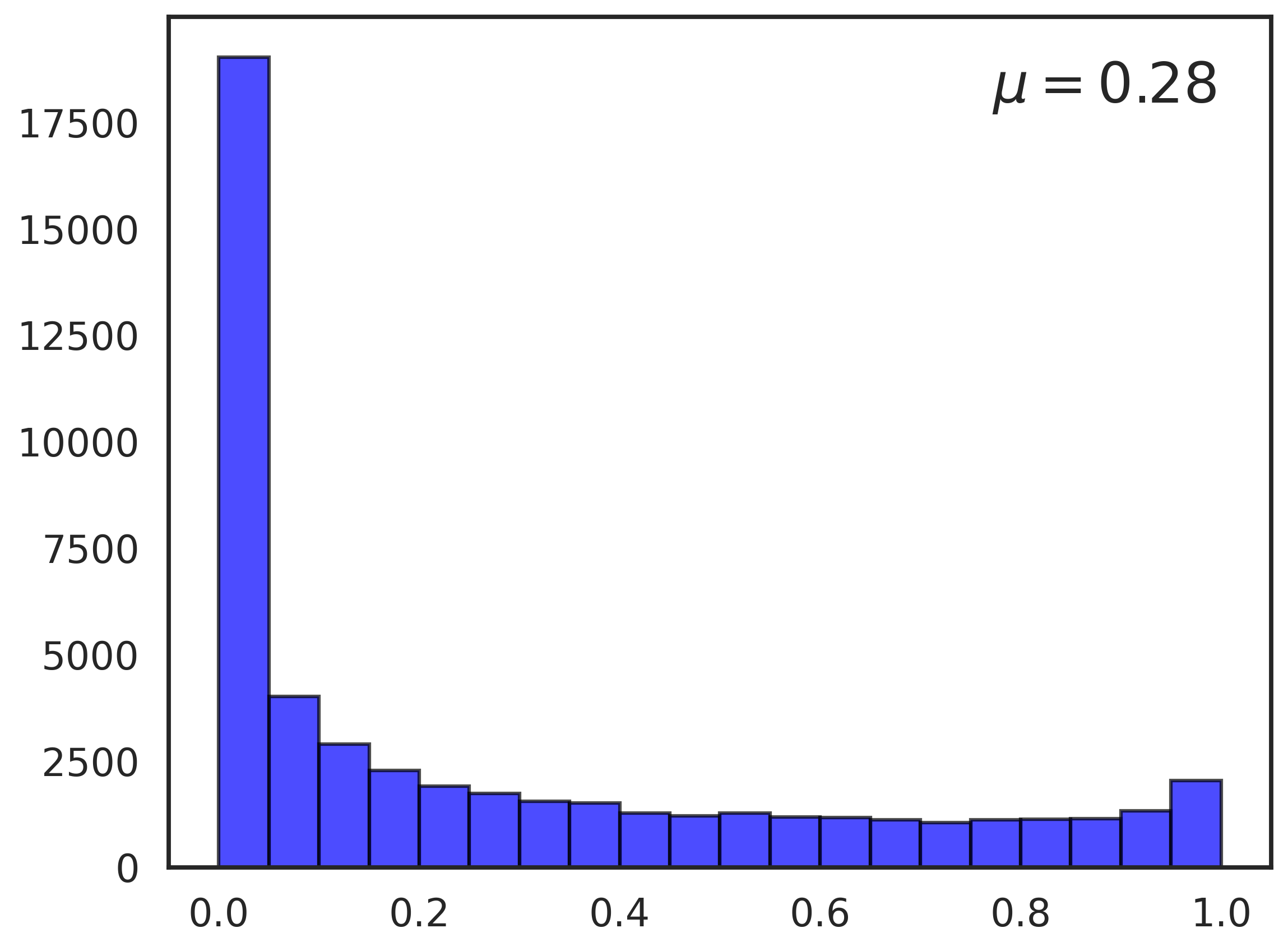}
    \subcaption{CIFAR-10, ResNet-18}
    \label{fig:memhist_c10_resnet}
  \end{subfigure}\hfill
  \begin{subfigure}[t]{0.32\textwidth}
    \centering
    \includegraphics[width=\textwidth]{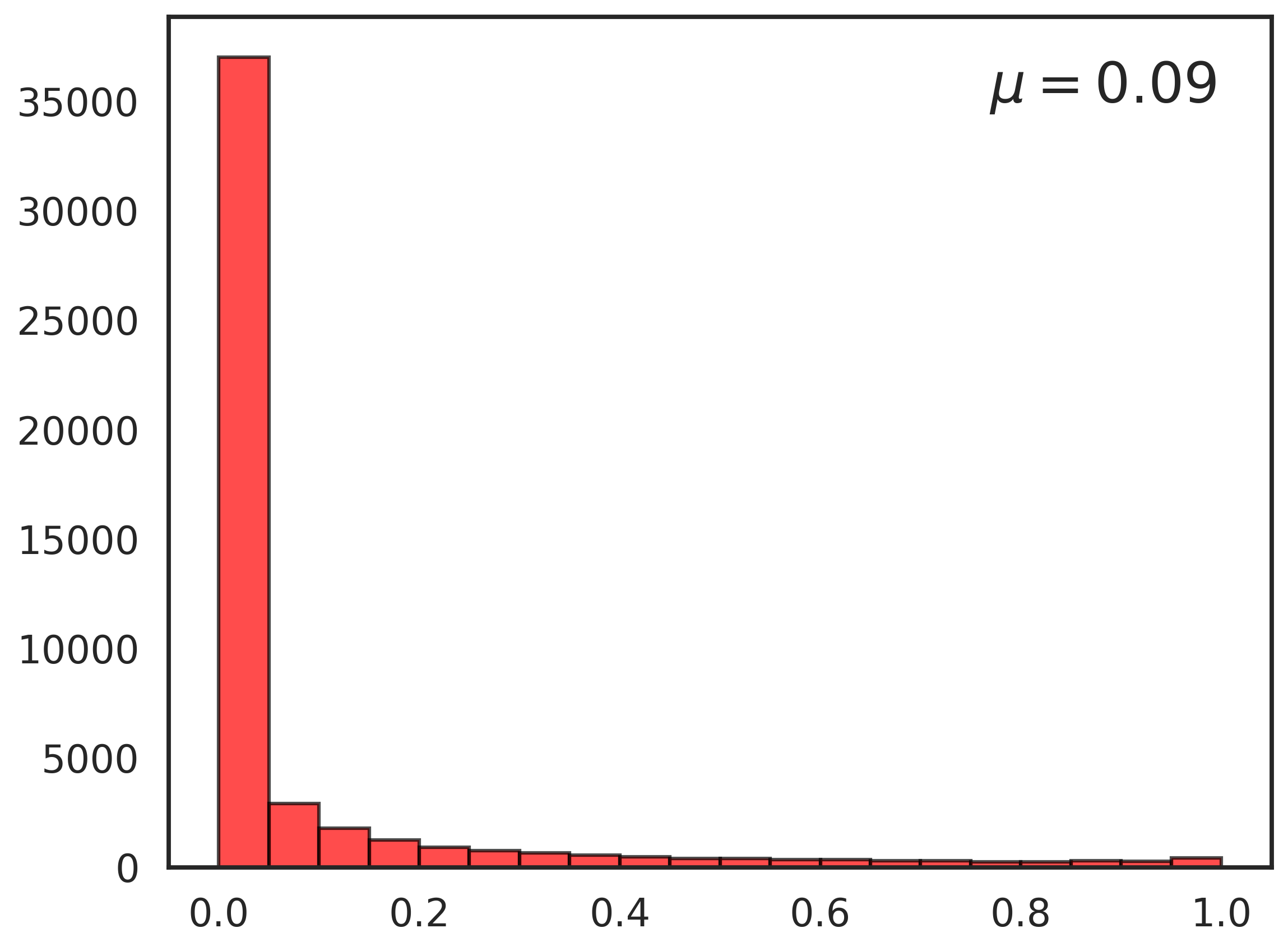}
    \subcaption{CIFAR-10, ViT-Small}
    \label{fig:memhist_c10_vit}
  \end{subfigure}\hfill
  \begin{subfigure}[t]{0.32\textwidth}
    \centering
    \includegraphics[width=\textwidth]{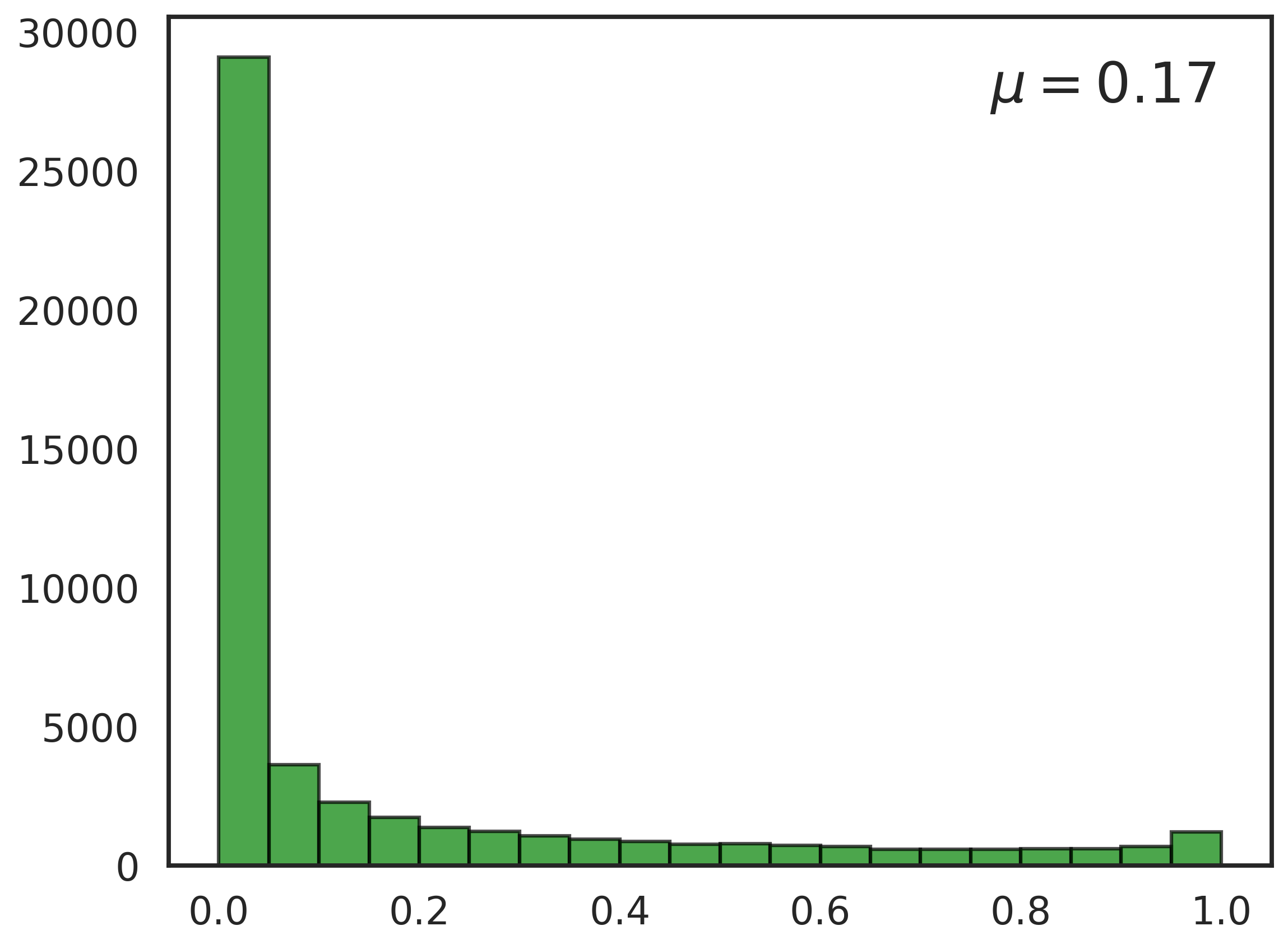}
    \subcaption{CIFAR-10, Swin-Tiny}
    \label{fig:memhist_c10_swin}
  \end{subfigure}

  \caption{
  Memorization histograms across architectures on CIFAR-100 (top) and CIFAR-10 (bottom). Across both datasets, the distributions are similarly skewed, consistent with the long-tail phenomenon in memorization.
  Memorization distributions on CIFAR-100 (top) and CIFAR-10 (bottom) for ResNet-50, ViT-Small, and Swin-Tiny. $\mu$ denotes the mean memorization score. 
All three architectures show a skewed, long-tailed distribution, suggesting that VTs, like CNNs, contain a small subset of highly memorized examples. This supports the use of memorization-based unlearning for VTs.}
  \label{fig:memorization_hist_both}
\end{figure*}

Figures \ref{fig:memorization_hist_both} 
present the memorization distributions across different architectures for CIFAR-100 and CIFAR-10. In all cases, we see similarly skewed distributions, consistent with the ``long-tail'' discovery in \cite{feldmanest}. 
For CIFAR-100, in particular, both VTs closely align with ResNet-50. 
Hence, despite architectural differences between VTs and CNNs, their memorization patterns remain fundamentally similar, especially for more complex tasks. This provides a  foundation for studying unlearning à la RUM in VTs.


For CIFAR-100, we observe that both VTs closely align with ResNet-50 in terms of memorization. However, on CIFAR-10, VTs exhibit slightly lower memorization compared to ResNet-18. 
This can be attributed to the fact that pretrained VTs benefit from a stronger ability to capture global contextual information, which reduces their reliance on memorizing training examples when adapting to simpler tasks. This advantage, however, does not extend to more challenging tasks like CIFAR-100.

\subsubsection{Are CNN-Derived Memorization Proxies Relevant for VTs?}
\label{subsec:proxy_corr}
Differences in architecture (e.g., inductive biases) and in training (e.g., finetune-then-retrain in VTs) raise doubts about the appropriateness of CNN memorization proxies for VTs. 

We first provide the formulas for computing each of the following proxies: Confidence (C), Max Confidence (MaxC), Entropy (E), Binary Accuracy (BA) \cite{jiang2021characterizingstructuralregularitieslabeled} and Holdout Retraining (HR) \cite{carlini2019distributiondensitytailsoutliers}. 

\begin{subequations}\label{eq:mem-proxies}
\setlength{\jot}{2pt} 
\begin{align}
\mathrm{Conf}(x_i,y_i) &= \frac{1}{E}\sum_{e=1}^E P_{\theta_e}(y=y_i \mid x_i), \\
\mathrm{MaxConf}(x_i) &= \frac{1}{E}\sum_{e=1}^E \max_{y} P_{\theta_e}(y \mid x_i), \\
\mathrm{Ent}(x_i) &= \frac{1}{E}\sum_{e=1}^{E}\sum_{y} P_{\theta_e}(y \mid x_i)\log P_{\theta_e}(y \mid x_i), \\
\mathrm{BA}(x_i,y_i) &= \frac{1}{E}\sum_{e=1}^{E} \mathds{1}\!\left[\arg\max_{y} P_{\theta_e}(y \mid x_i)=y_i\right], \\
\mathrm{HR}(x_i) &= \mathrm{symKL}\!\big(p_{\theta_o}(x_i)\,\|\,p_{\theta'}(x_i)\big).
\end{align}
\end{subequations}

Results in Table \ref{tab:spearman_correlations} answer the question in the affirmative. Proxies were calculated using models trained with identical hyper-parameters as those used to derive memorization scores (see Table \ref{tab:train_config_est} in Appendix \ref{app:memest}). 

\begin{table}[h]
\centering
\caption{Spearman correlation coefficients between memorization and proxies (duplicate of Table \ref{tab:spearman_correlations}, included here for completeness)}
    \begin{tabular}{lccccccc}
        \toprule
        & \multicolumn{3}{c}{\textbf{CIFAR-10}} & & \multicolumn{3}{c}{\textbf{CIFAR-100}} \\
        \cmidrule{2-4} \cmidrule{6-8}
        \textbf{Proxy} & \textbf{ResNet-18} & \textbf{ViT-Small} & \textbf{Swin-Tiny} & & \textbf{ResNet-50} & \textbf{ViT-Small} & \textbf{Swin-Tiny} \\
        \midrule
        \textbf{Conf}          & -0.80 & -0.79 & -0.88 & & -0.91 & -0.85 & -0.90 \\
        MaxConf      & -0.76 & -0.77 & -0.85 & & -0.87 & -0.80 & -0.86 \\
        Ent             & -0.75 & -0.78 & -0.85 & & -0.80 & -0.77 & -0.82 \\
        BA     & -0.71 & -0.63 & -0.79 & & -0.89 & -0.69 & -0.78 \\
        \textbf{HR}  & +0.67  & +0.45  & +0.64  & & +0.62  & +0.50  & +0.52  \\
        \bottomrule
    \end{tabular}
\end{table}
Confidence consistently demonstrates the strongest correlation across all models and datasets. 
The correlation magnitude for VTs largely resembles that of CNNs -- further evidence that memorizations are fundamentally similar across these architectures. Swin-Tiny generally has stronger correlations compared to ViT-Small, possibly due to its hierarchical structure that more closely resembles traditional CNNs. 
Holdout Retraining shows more moderate positive (albeit still significant) correlations.

Based on these findings, we selected 
Confidence as the best-performing learning event proxy with consistently high absolute correlation values and Holdout Retraining as a representative of a different proxy category. Importantly, Holdout Retraining offers large computational advantages as it does not require monitoring model behaviour during training.

\subsection{Representation-level Analysis}
\label{app:representation_analysis}

\subsubsection{Embedding Visualization of Forget vs. Retain Behavior}

\begin{figure}[htb]
  \centering
    \begin{subfigure}{0.49\textwidth}
      \centering
      \includegraphics[width=0.9\columnwidth]{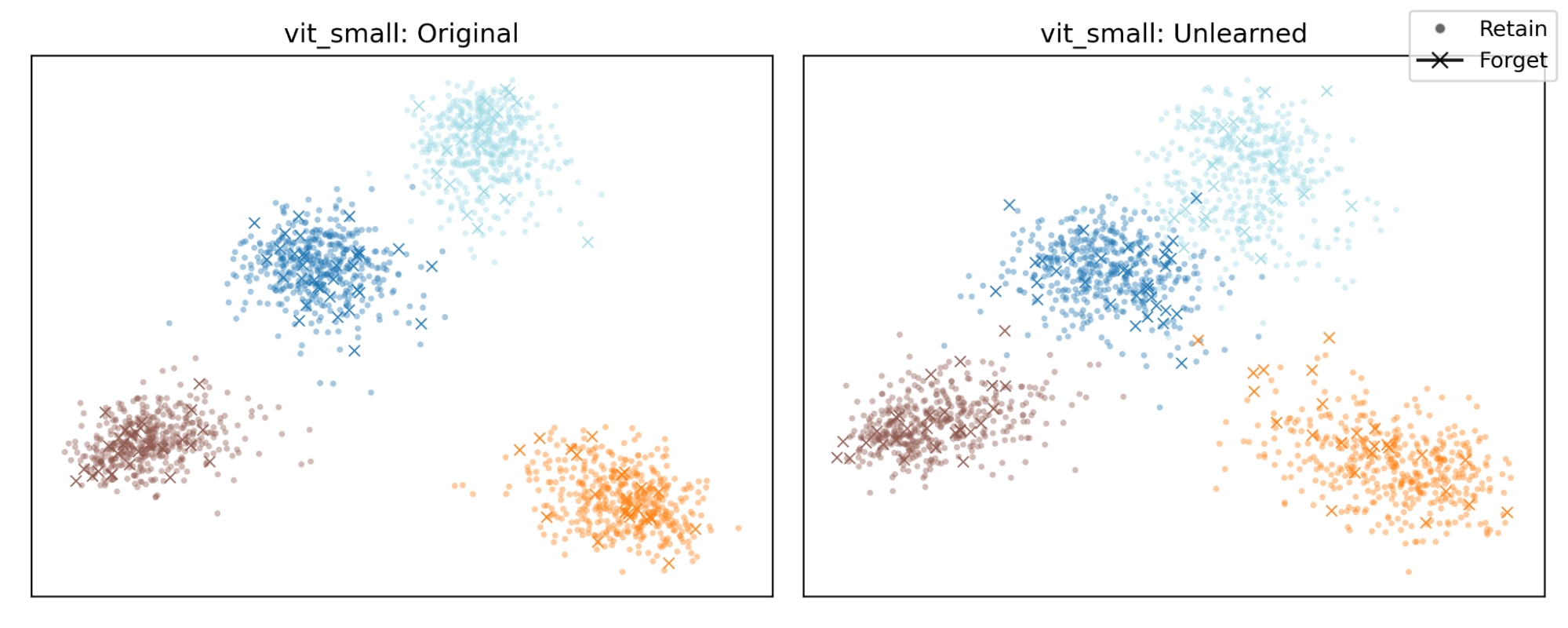}
      \subcaption{ViT-Small}
    \end{subfigure}
    \begin{subfigure}[b]{0.49\textwidth}
      \centering
      \includegraphics[width=0.9\columnwidth]{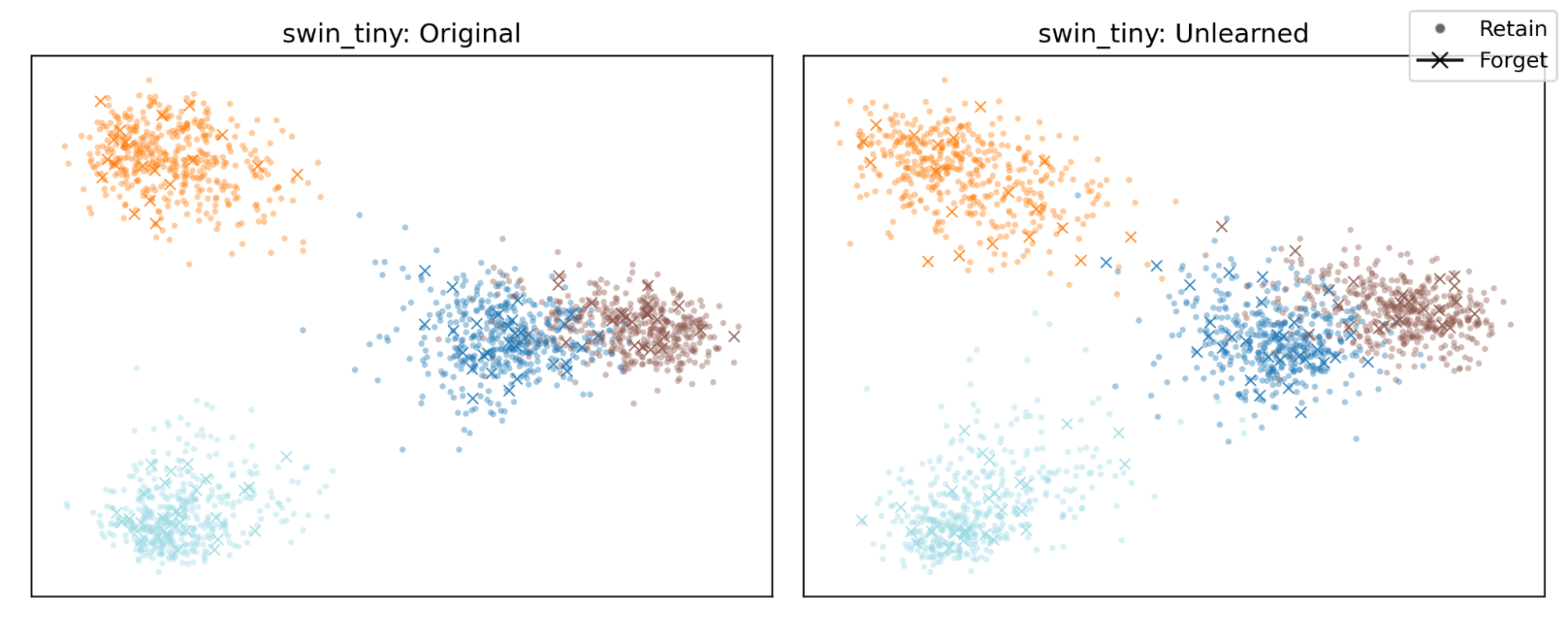}
      \subcaption{Swin-Tiny}
    \end{subfigure}
  \caption{Representation visualization before vs after unlearning for ViT-Small and Swin-Tiny}
  \label{fig:visualization}
\end{figure}

To complement our quantitative results, we include qualitative visualizations based on 2-D PCA projections of the models’ embedding space, as shown in Figure \ref{fig:visualization}.

For each architecture (ViT-Small and Swin-Tiny), and for the same set of samples, we extract image-level embeddings from:
the original (pre-unlearning) model, and
the unlearned model (NegGrad+ with the HR proxy),
and then project both sets of embeddings into two dimensions using PCA.
We visualize four example classes. Within each class, retain examples are shown as dots, and forget examples are shown as crosses of the same color. The left panel of each figure shows the original model, and the right panel shows the unlearned model.

We can see from Figure \ref{fig:visualization} that, before unlearning, clusters are more compact, with forget examples sitting within the clusters. After unlearning, we see clearly that: (i) forget examples are “being pushed out” of the core cluster; this is a strong indication that confidence of prediction is weakened for forget examples, which aligns with our quantitative Forget Accuracy and MIA results. (ii) remain examples stay largely within their core clusters, which aligns with our quantitative Retain Accuracy results.

\subsubsection{Layer-wise CKA analysis.}
To further examine how unlearning changes internal representations, we compute layer-wise Centered Kernel Alignment (CKA) between representations from the original model $\theta_o$, the retrained reference model $\theta_r$, and the unlearned model $\theta_u$. We compare three model pairs: original--retrain $(\theta_o,\theta_r)$, unlearned--original $(\theta_u,\theta_o)$, and unlearned--retrain $(\theta_u,\theta_r)$. Higher CKA indicates more similar representations at a given layer.

\begin{figure}[htb]
  \centering
    \begin{subfigure}{0.49\textwidth}
      \centering
      \includegraphics[width=0.9\columnwidth]{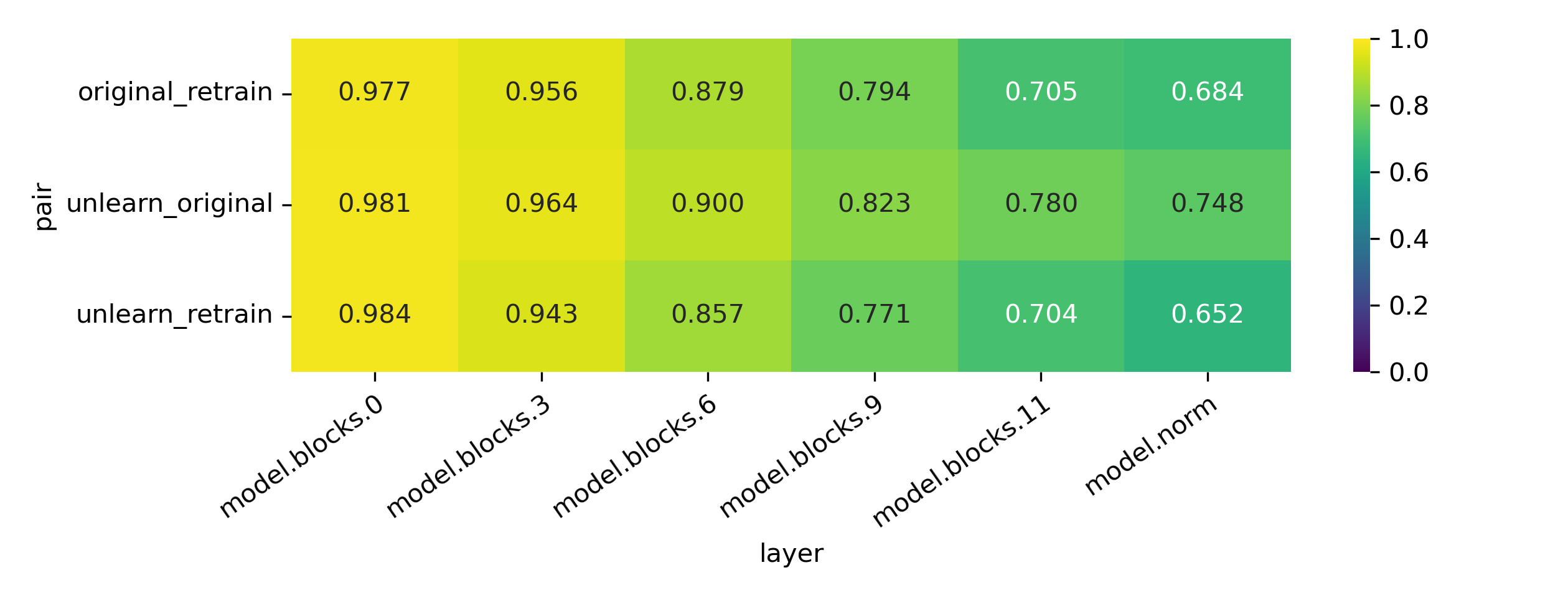}
      \subcaption{ViT-Small, Fine-tune}
    \end{subfigure}
    \begin{subfigure}[b]{0.49\textwidth}
      \centering
      \includegraphics[width=0.9\columnwidth]{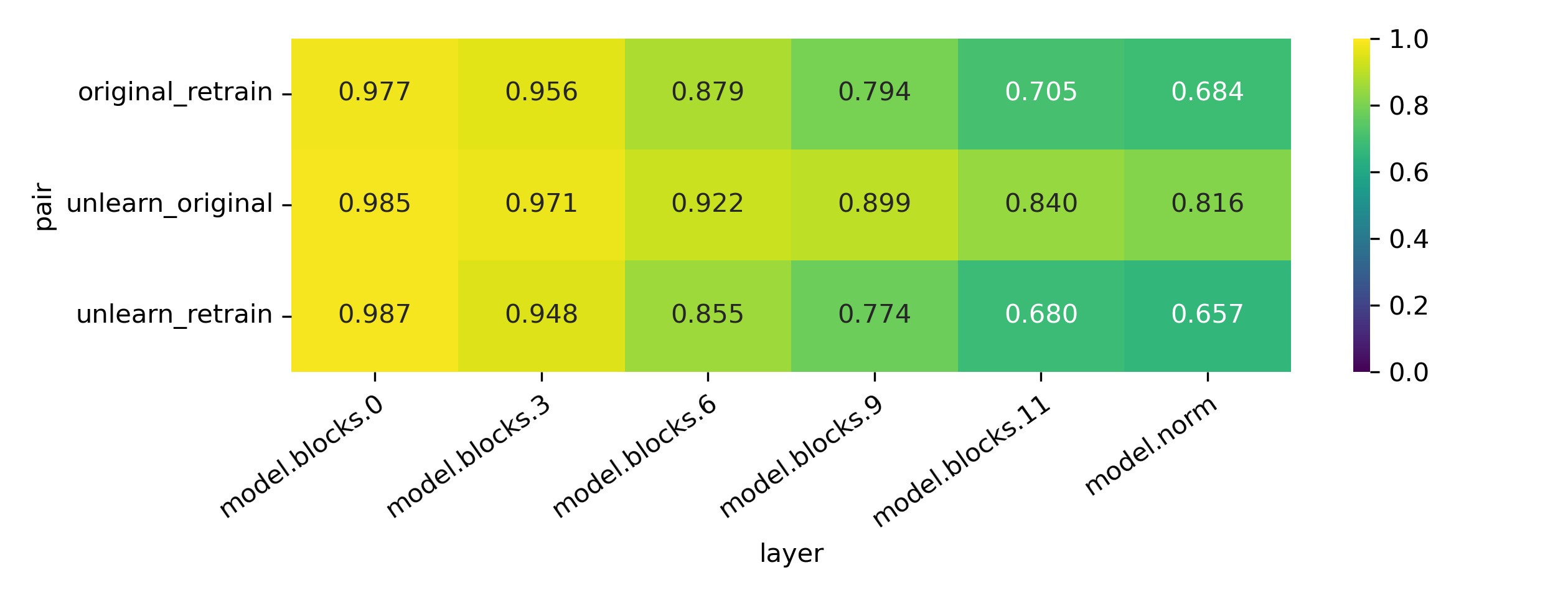}
      \subcaption{ViT-Small, NegGrad+}
    \end{subfigure}

  \centering
    \begin{subfigure}{0.49\textwidth}
      \centering
      \includegraphics[width=0.8\columnwidth]{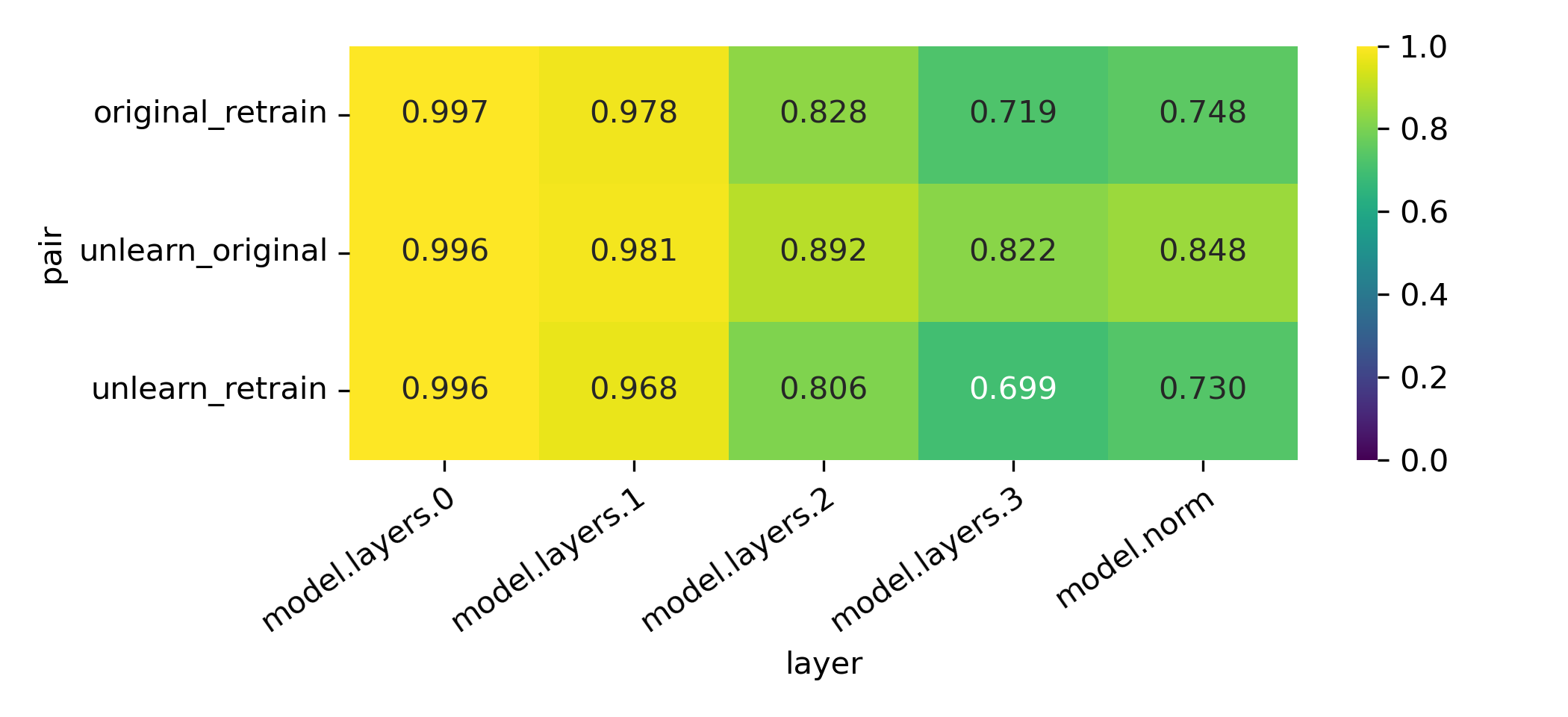}
      \subcaption{Swin-Tiny, Fine-tune}
    \end{subfigure}
    \begin{subfigure}[b]{0.49\textwidth}
      \centering
      \includegraphics[width=0.8\columnwidth]{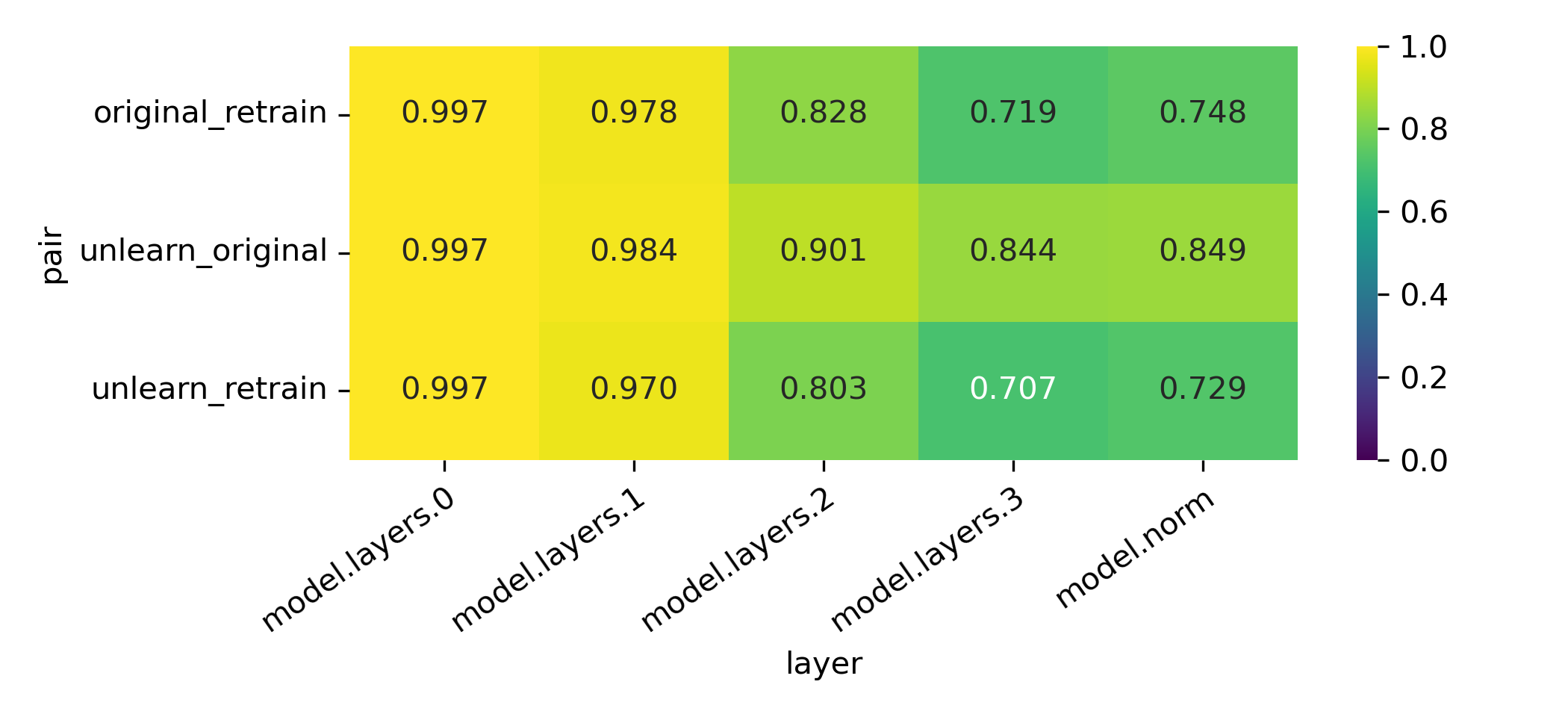}
      \subcaption{Swin-Tiny, NegGrad+}
    \end{subfigure}
  \caption{
  Layer-wise CKA similarity on CIFAR-100 for ViT-Small and Swin-Tiny under Fine-tune and NegGrad+. Rows compare model pairs, columns correspond to layers, and higher values indicate more similar representations. Early layers remain highly stable, while larger changes occur in later blocks and the final normalization layer. 
  }
  \label{fig:cka_analysis}
\end{figure}

Figure \ref{fig:cka_analysis} shows representative CKA heatmaps on CIFAR-100 for ViT-Small and Swin-Tiny under Fine-tune and NegGrad+. Across architectures and methods, early layers remain highly similar across model pairs, while larger representation changes appear in later blocks and the final normalization layer. This suggests that both retraining and unlearning mainly affect higher-level representations, while early feature extraction remains largely stable.

We also observe architecture-specific patterns. ViT-Small shows sharper sensitivity in later layers, especially near the final transformer blocks and normalization layer. In contrast, Swin-Tiny exhibits a more staged pattern of representation change, consistent with its hierarchical design. These observations provide complementary evidence that architecture-specific unlearning behavior is most visible in higher-level representations.

Interestingly, although $\theta_r$ is the functional target for unlearning, $\theta_u$ does not necessarily become more similar to $\theta_r$ than $\theta_o$ is in layer-wise representation space. This indicates that successful unlearning does not require the unlearned model to reproduce the internal representation trajectory of full retraining. Instead, $\theta_u$ can achieve strong ToW and ToW-MIA performance while following a different representational path.

\subsection{Additional Dataset Results}
This section presents the corresponding figures for the CIFAR-10 dataset that mirror the ones of CIFAR-100 found in Section \ref{sec:exp_ana}. 

\subsubsection{Unlearning Results on CIFAR-10}
\label{app:unlearnperfcifar10}


\begin{figure*}[htb]
  \centering
  \begin{tabular}{@{}cc@{}}
    \begin{minipage}{0.48\textwidth}
      \centering
      \includegraphics[width=0.9\columnwidth]{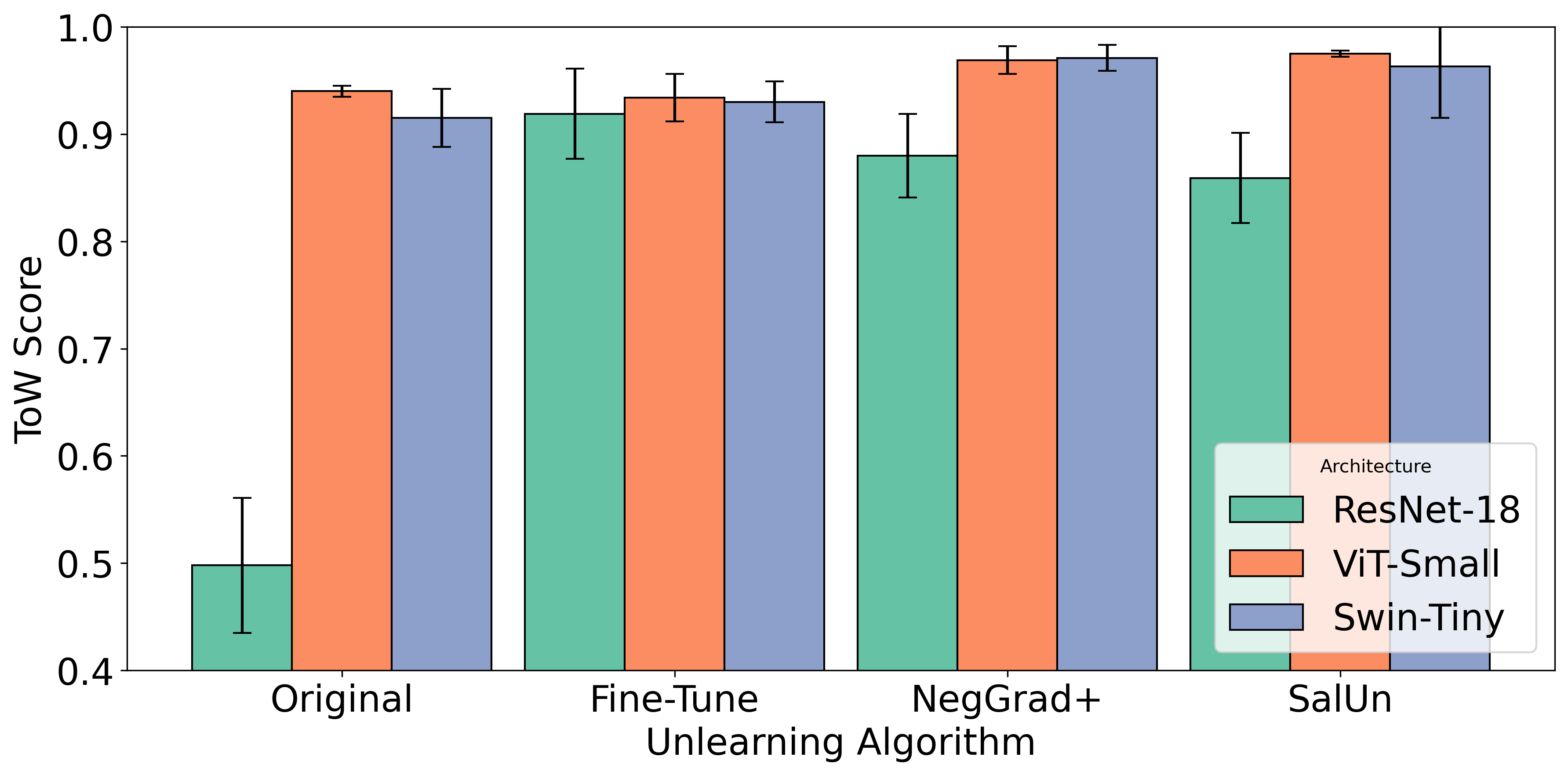}
      \subcaption{ToW with Confidence}
    \end{minipage} &
    \begin{minipage}{0.48\textwidth}
      \centering
      \includegraphics[width=0.9\columnwidth]{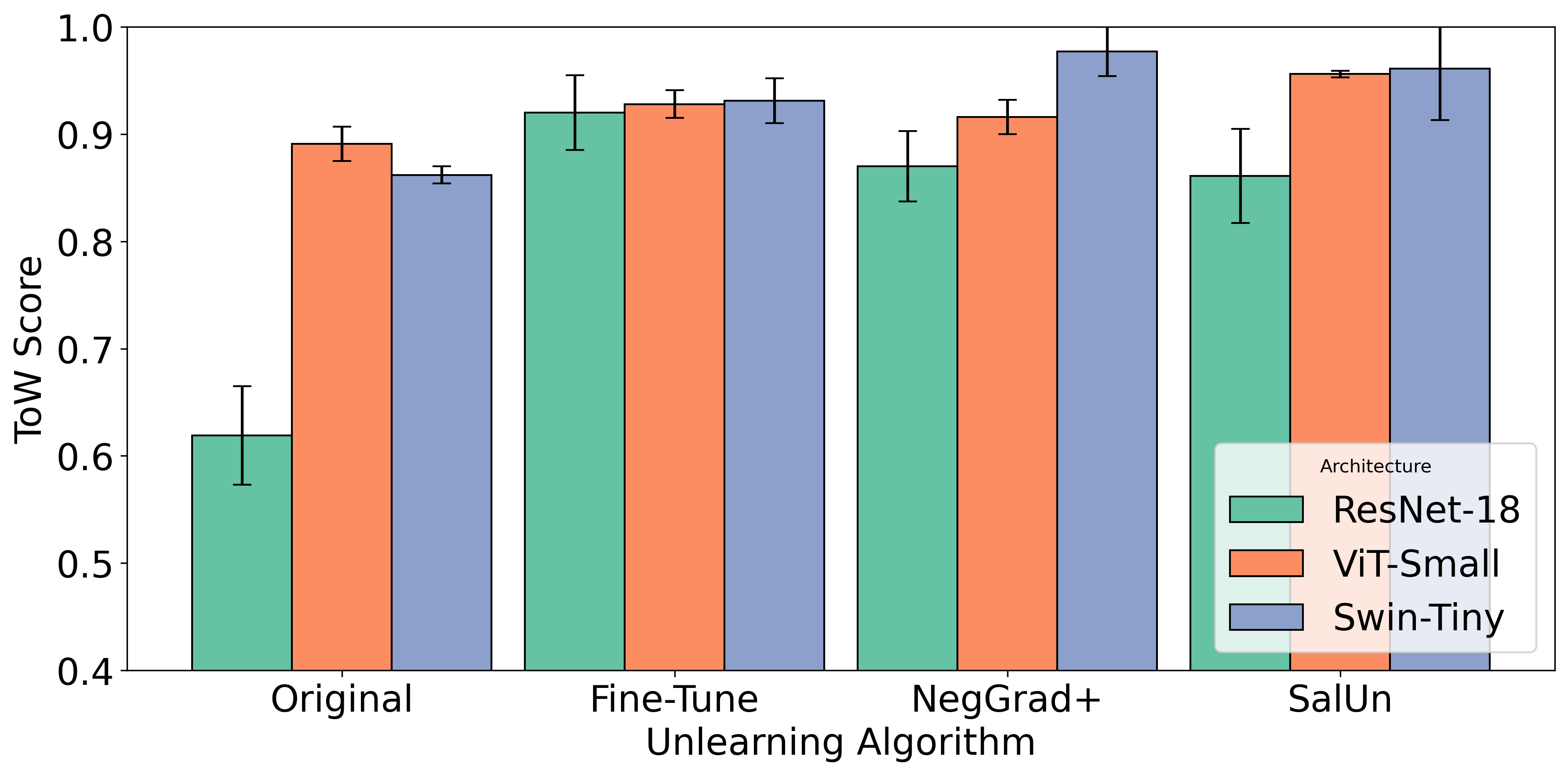}
      \subcaption{ToW with Holdout Retraining}
    \end{minipage} \\
    \begin{minipage}{0.48\textwidth}
      \centering
      \includegraphics[width=0.9\columnwidth]{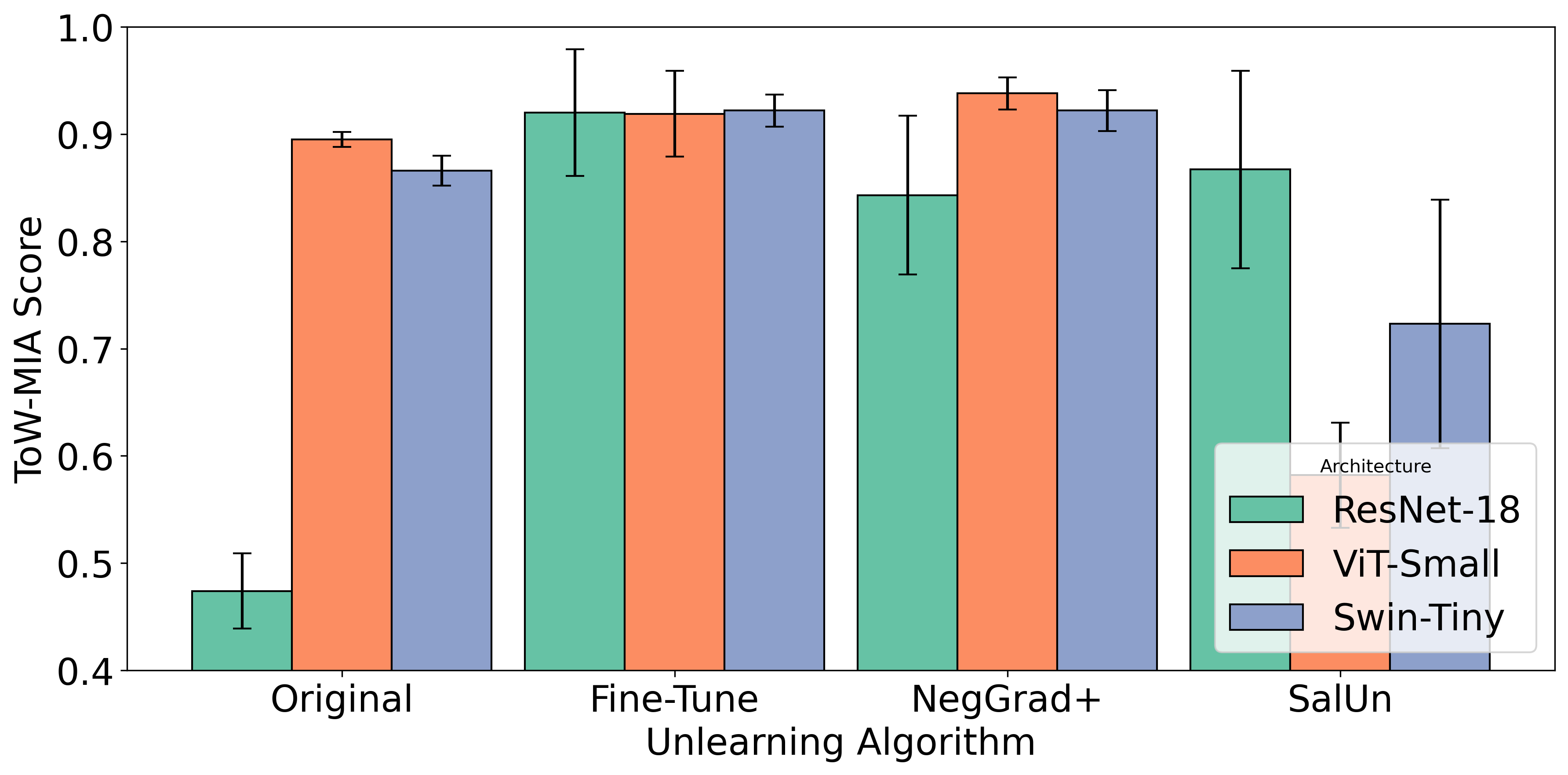}
      \subcaption{ToW-MIA with Confidence}
    \end{minipage} &
    \begin{minipage}{0.48\textwidth}
      \centering
      \includegraphics[width=0.9\columnwidth]{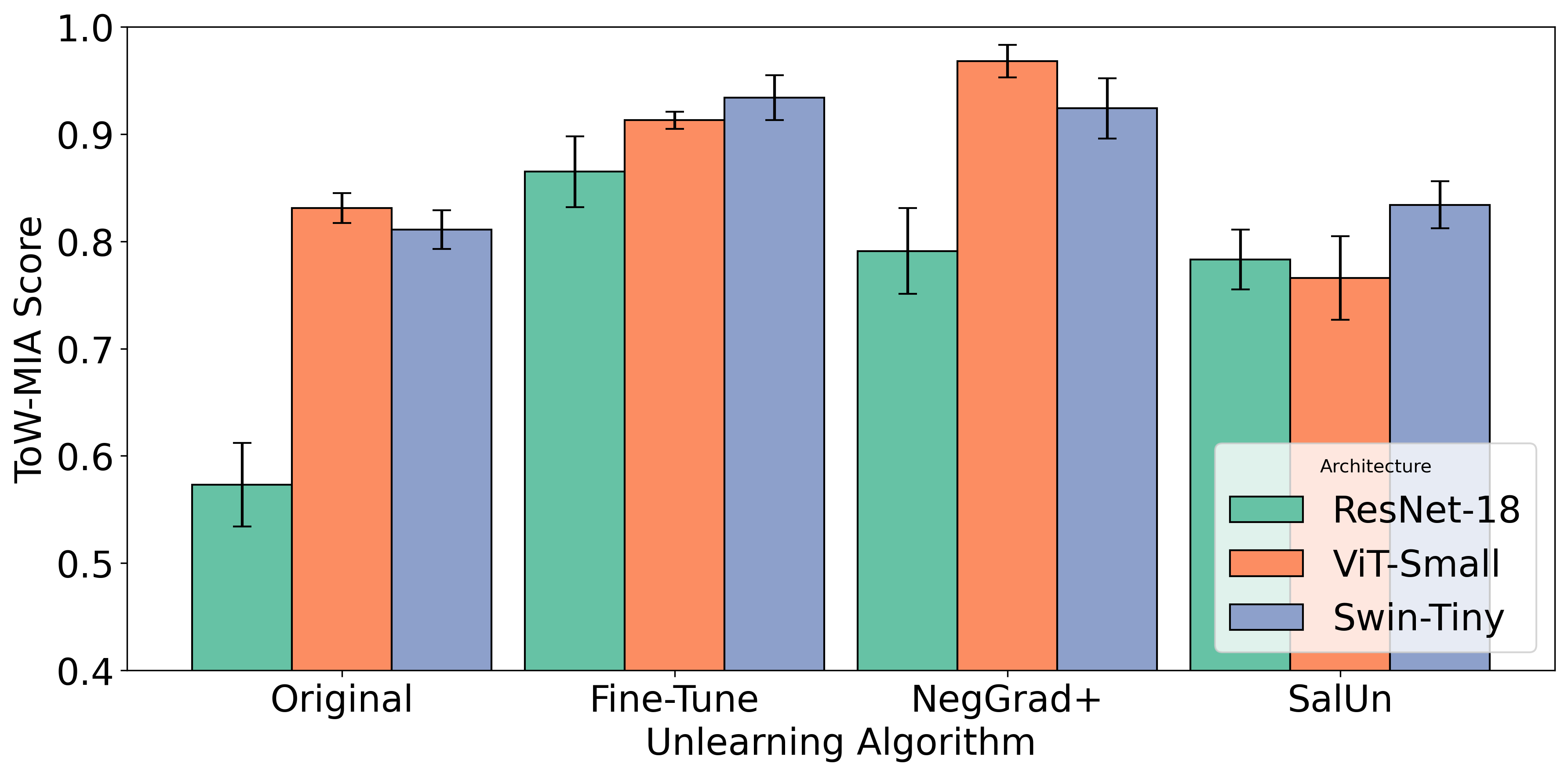}
      \subcaption{ToW-MIA with Holdout Retraining}
    \end{minipage} \\
  \end{tabular}
  \caption{Architecture performance comparison on CIFAR-10}
  \label{fig:cifar10_unlearning_comparison}
\end{figure*}

\textbf{Algorithm-Level Comparison on CIFAR-10.}
Figure \ref{fig:cifar10_unlearning_comparison} shows the unlearning performance results for the CIFAR-10 dataset.
First, note that the original VT model (no unlearning) already achieves relatively high scores on CIFAR-10, which is unlike the results in CIFAR-100. This is consistent with our earlier observation that VTs exhibit lower memorization on this simpler dataset, reducing performance differences between $\theta_o$ and $\theta_r$ on $D_f$. Nevertheless, we see marginal improvements from unlearning algorithms on CIFAR-10, still making a case for their utilization in low-memorized data.

Taking a closer look, all methods in VTs appear to show smaller improvements in ToW compared to the original model. 
Additionally, there is no clear winner for VTs between the two proxy strategies.
Again, these results are in contrast to the more substantial improvements observed on the more complex CIFAR-100 dataset. For instance, even a simple approach like Fine-tune significantly outperforms the original model across all configurations on CIFAR-100—an effect not observed here.
More advanced algorithms show greater gains in ToW performance on CIFAR-100 than they do on CIFAR-10, compared to the original VT model.

\textbf{Architecture-Level Comparison on CIFAR-10.} 
\label{app:archcifar10}
Figure \ref{fig:cifar10_unlearning_comparison} also presents the architecture-wise comparison on CIFAR-10, analogous to the analysis shown for CIFAR-100 in Figure \ref{fig:cifar100_comparison}. We observe that VTs generally achieve higher ToW scores than ResNet-18, with NegGrad+ and SalUn performing particularly well on both VTs. 
However, this performance gap is less evident in the ToW-MIA metrics, where Fine-tune shows comparable performance across all architectures. 
Notably, NegGrad+ is the only unlearning method that improves ToW-MIA for both proxies. In contrast, SalUn continues to struggle on ToW-MIA within VTs:
ResNet-18 significantly outperforms VTs when using the Confidence proxy and matches their performance under Holdout Retraining.

\subsubsection{Unlearning Results on SVHN}
\begin{figure*}[h]
  \centering
    \begin{subfigure}{0.48\textwidth}
      \centering
      \includegraphics[width=0.9\columnwidth]{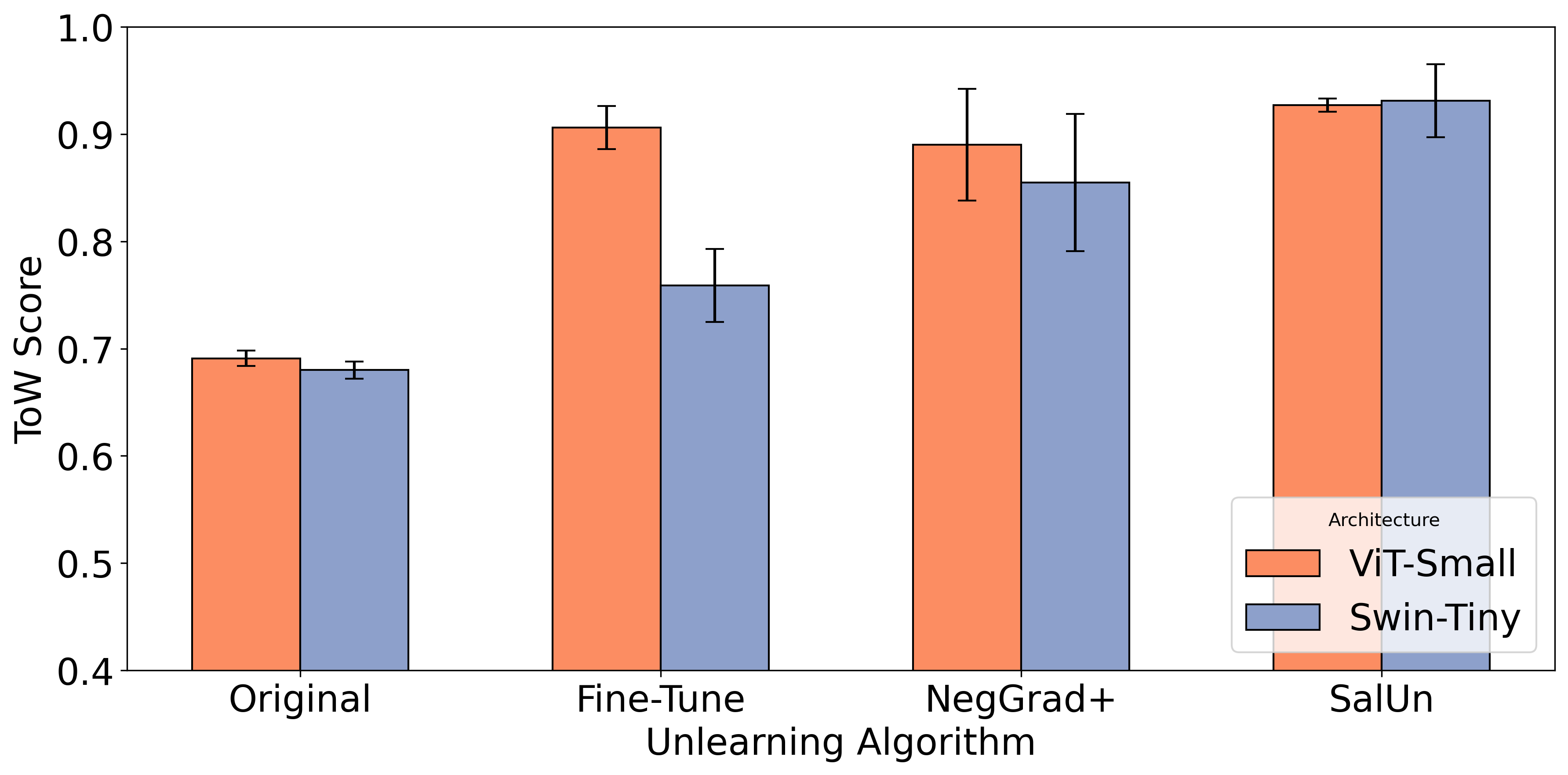}
      \caption{ToW scores with Confidence}
    \end{subfigure} 
    \begin{subfigure}{0.48\textwidth}
      \centering
      \includegraphics[width=0.9\columnwidth]{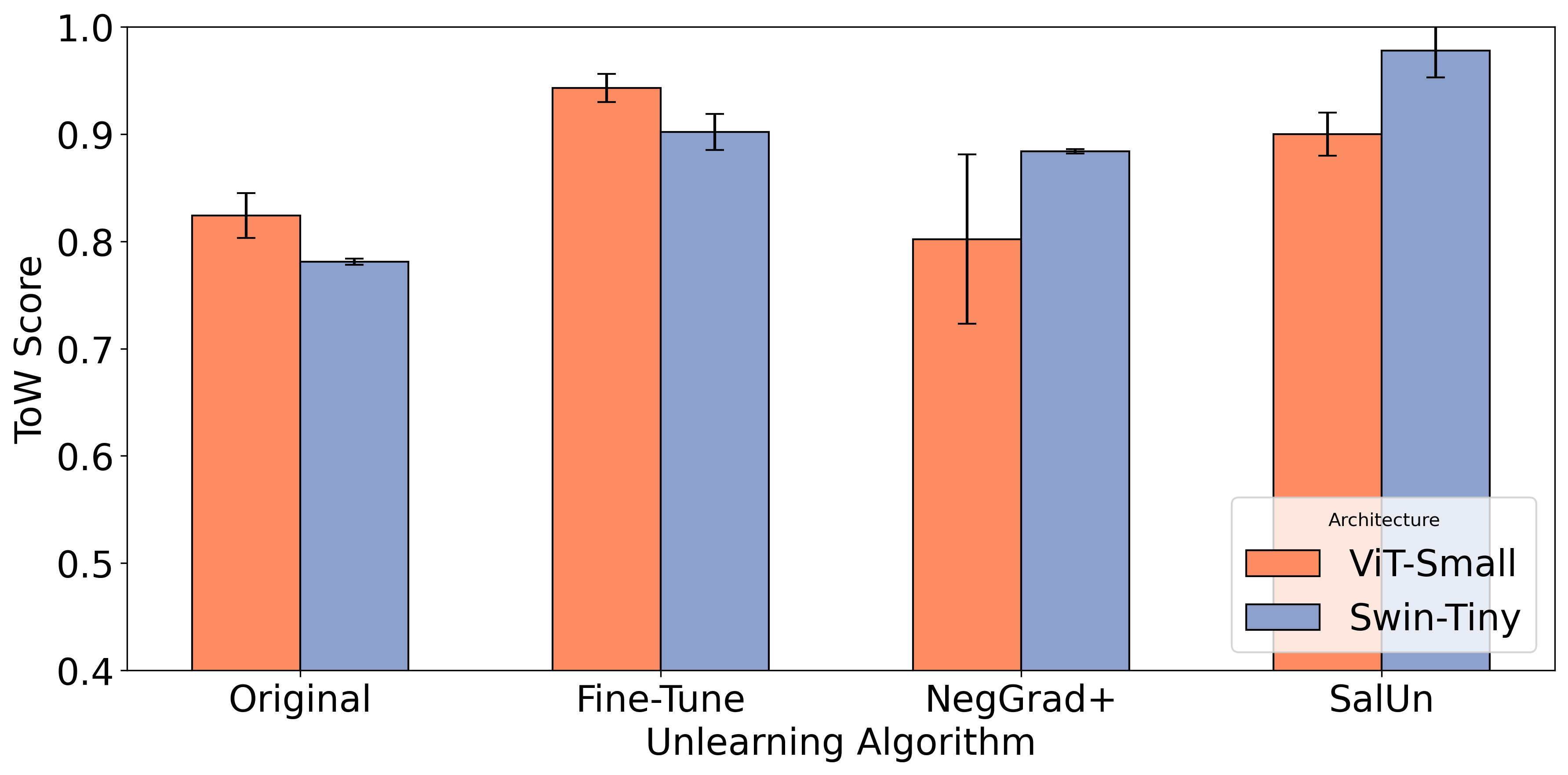}
      \caption{ToW scores with Holdout Retraining}
    \end{subfigure} \\
    \begin{subfigure}{0.48\textwidth}
      \centering
      \includegraphics[width=0.9\columnwidth]{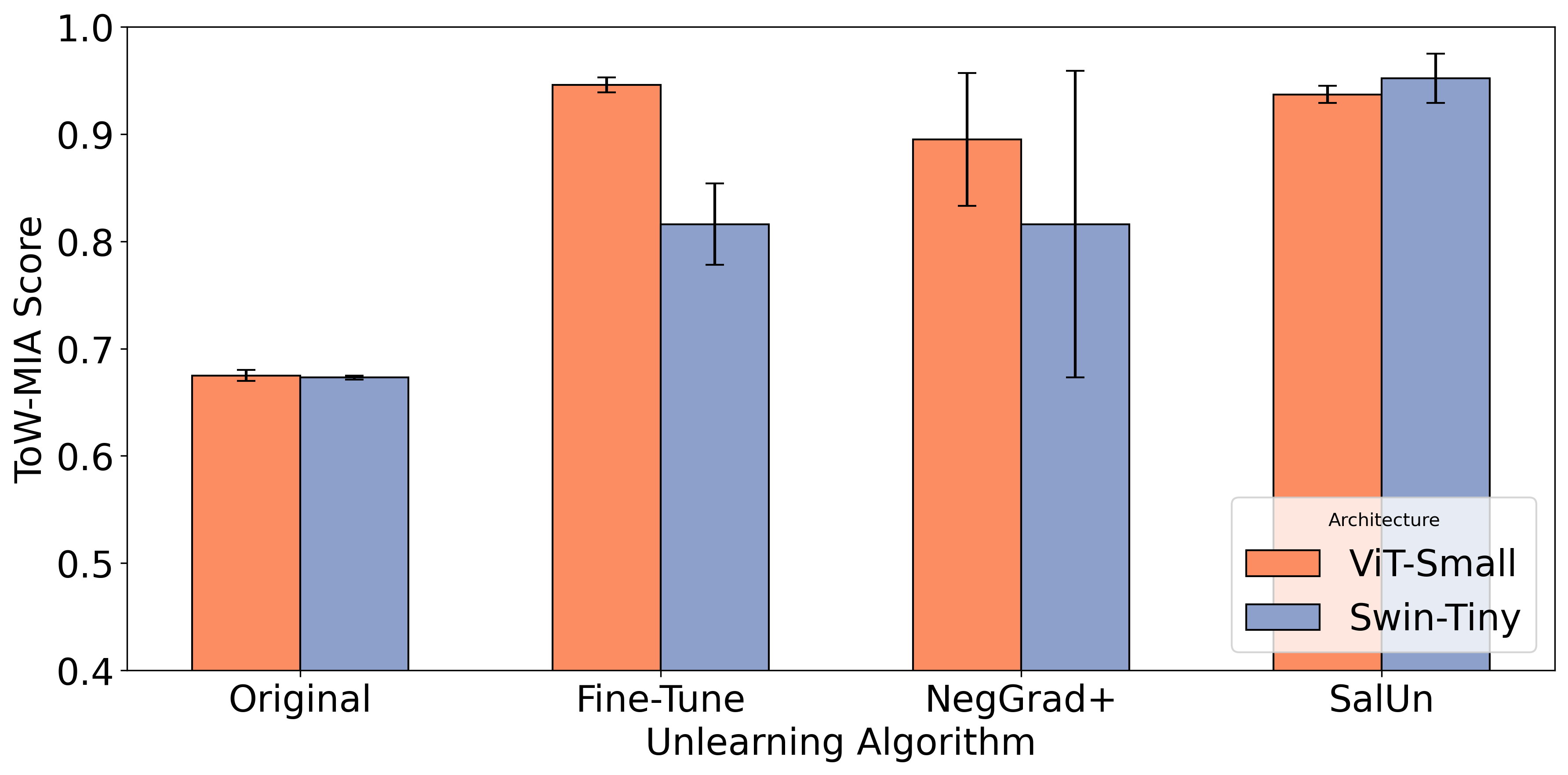}
      \caption{ToW-MIA scores with Confidence}
    \end{subfigure} 
    \begin{subfigure}{0.48\textwidth}
      \centering
      \includegraphics[width=0.9\columnwidth]{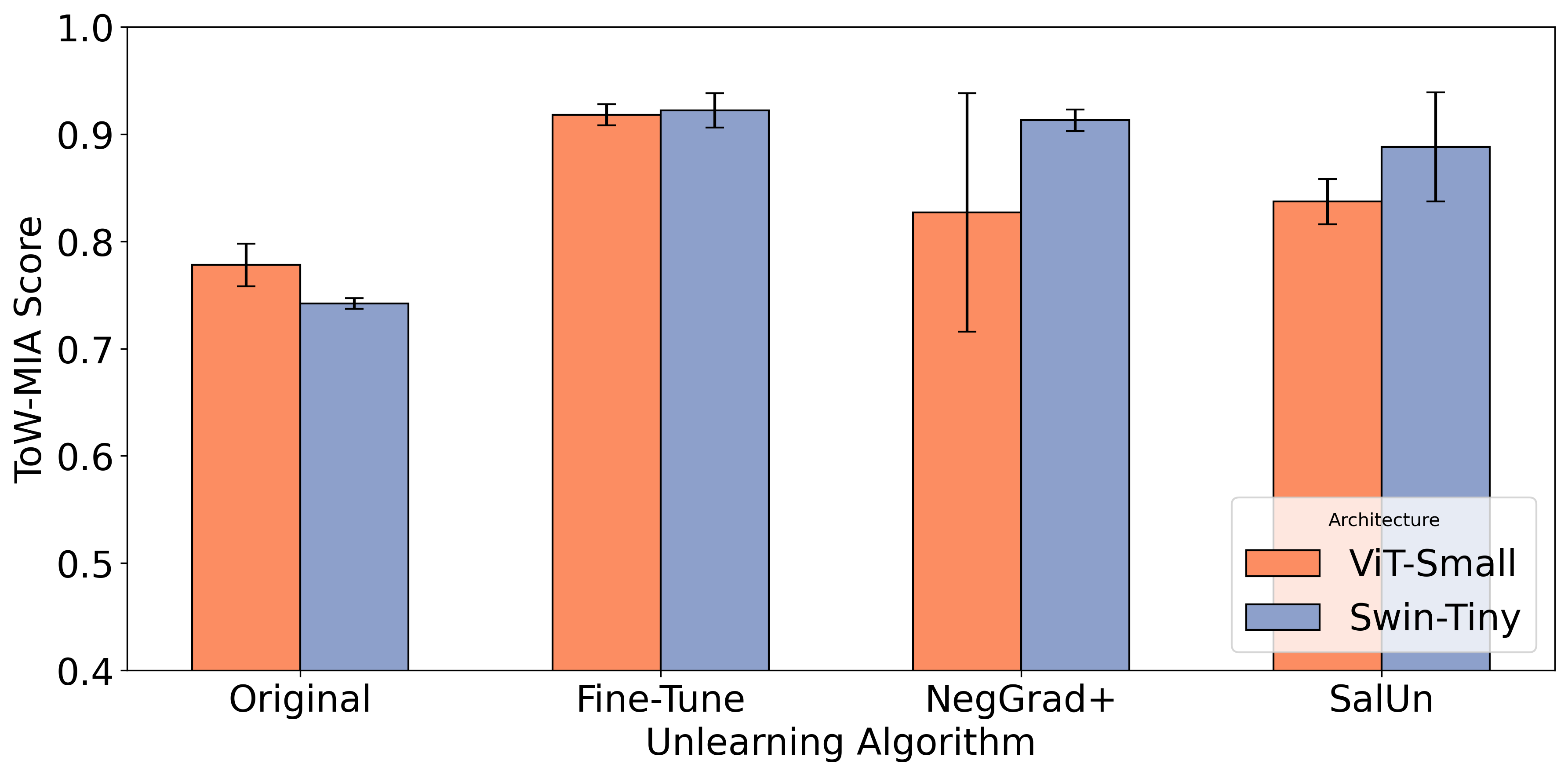}
      \caption{ToW-MIA scores with Holdout Retraining}
    \end{subfigure}
  \caption{MU performance comparison on SVHN}
  \label{fig:svhn_unlearning_comparison}
\end{figure*}

To extend our study to a different dataset, we repeat the experiments on SVHN using both ViT-Small and Swin-Tiny architectures. Table \ref{tab:train_config_train} lists the hyperparameter settings used for the original model training and subsequent retraining.
Figure \ref{fig:svhn_unlearning_comparison} shows the comparative unlearning performance of these methods on SVHN. 
See Section \ref{subsec:def_unlearning} for a detailed discussion of the results.


\subsection{Continual Unlearning Results}
\label{app:continual_mu_cifar10}

\begin{figure}[htb]
  \centering
      \begin{subfigure}{0.45\textwidth}
      \includegraphics[width=0.9\columnwidth]{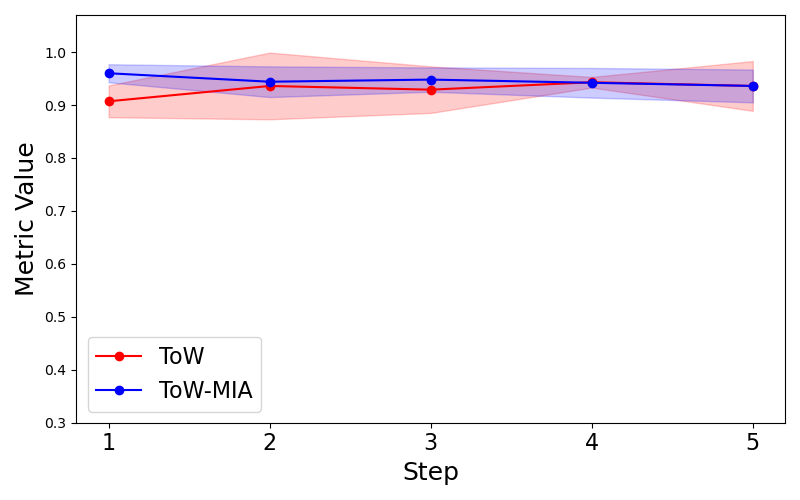}
      \subcaption{ViT-Small, CIFAR-10}
      \label{fig:lineplot_cifar10_vit_small}
    \end{subfigure}
    \begin{subfigure}{0.45\textwidth}
      \includegraphics[width=0.9\columnwidth]{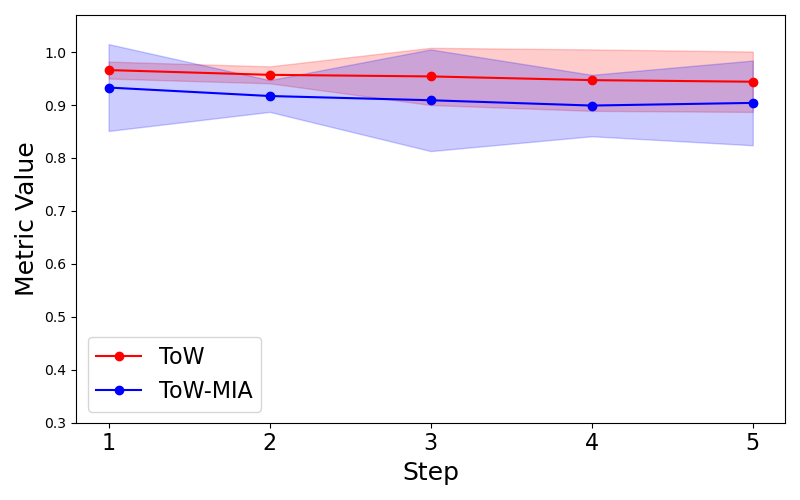}
      \subcaption{Swin-Tiny, CIFAR-10}
      \label{fig:lineplot_cifar10_swin_tiny}
    \end{subfigure}
    
    \begin{subfigure}{0.45\textwidth}
      \includegraphics[width=0.9\columnwidth]{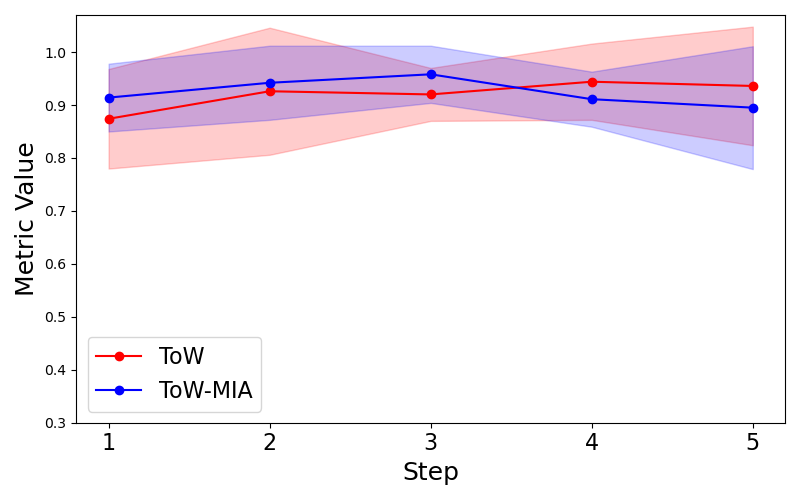}
      \subcaption{ViT-Small, SVHN}
      \label{fig:lineplot_svhn_vit_small}
    \end{subfigure}
    \begin{subfigure}{0.45\textwidth}
      \includegraphics[width=0.9\columnwidth]{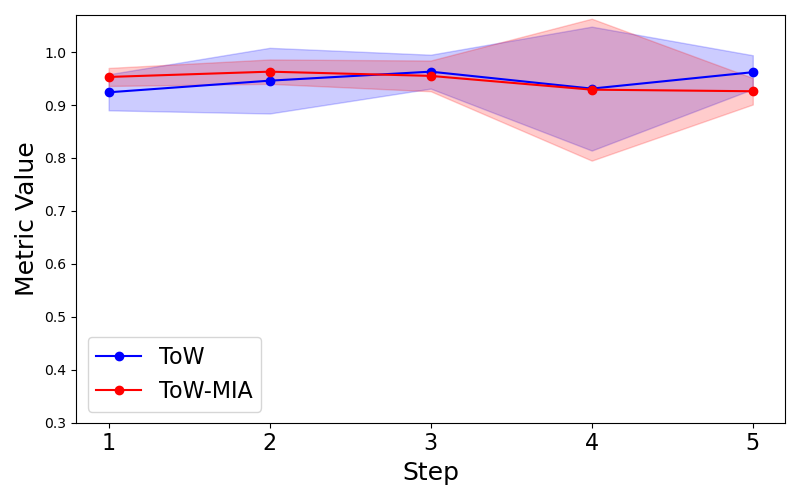}
      \subcaption{Swin-Tiny, SVHN}
      \label{fig:lineplot_svhn_swin_tiny}
    \end{subfigure}
  \caption{Continual unlearning performance across 5 steps on CIFAR-10 and SVHN datasets using NegGrad+ and the HR proxy.}
  \label{fig:sequential_unlearning_cifar10}
\end{figure}

\begin{figure}[htb]
  \centering
    \begin{subfigure}{0.45\textwidth}
      \includegraphics[width=0.9\columnwidth]{lineplots/seq_vit_small_cifar100.png}
      \subcaption{ViT-Small, 5 steps}
    \end{subfigure}
    \begin{subfigure}{0.45\textwidth}
      \includegraphics[width=0.9\columnwidth]{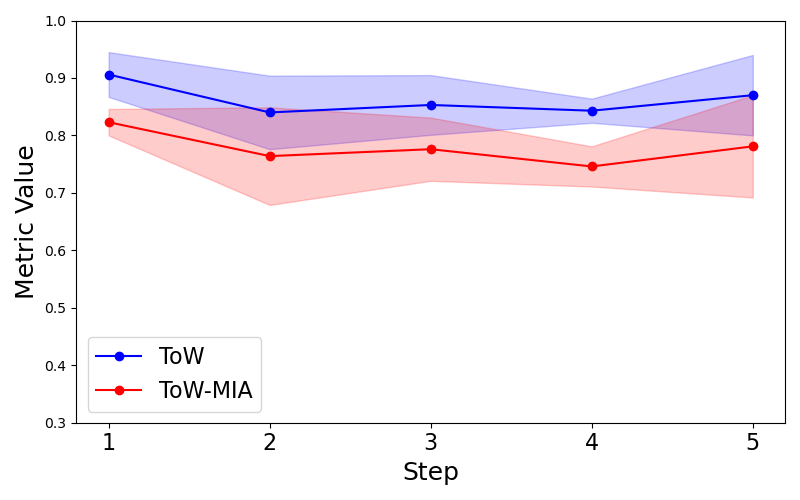}
      \subcaption{Swin-Tiny, 5 steps}
    \end{subfigure}
    \begin{subfigure}{0.45\textwidth}
      \includegraphics[width=0.9\columnwidth]{lineplots/seq_vit_small_cifar100_10steps.png}
      \subcaption{ViT-Small, 10 steps}
    \end{subfigure}
    \begin{subfigure}{0.45\textwidth}
      \includegraphics[width=0.9\columnwidth]{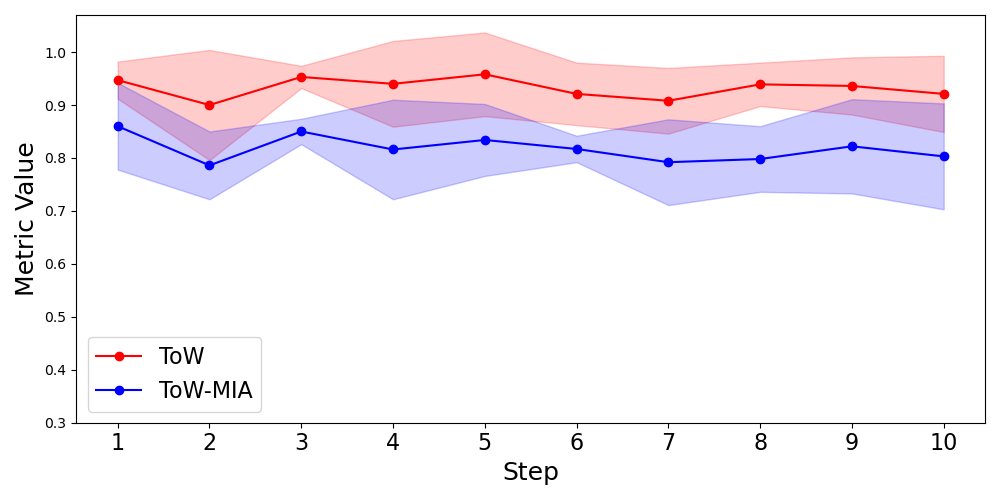}
      \subcaption{Swin-Tiny, 10 steps}
    \end{subfigure}
  \caption{Continual unlearning performance across 5/10 steps on CIFAR-100 using NegGrad+ and the HR proxy.}
  \label{fig:sequential_unlearning_10steps}
\end{figure}

We follow the same procedure 
outlined in Section \ref{subsec:stab_prox}
to sequentially apply NegGrad+ and Holdout Retraining on the CIFAR-10, CIFAR-100, and SVHN datasets over five consecutive steps.
The results are presented in Figure \ref{fig:sequential_unlearning_cifar10}, evaluated by ToW and ToW-MIA. 
Figure \ref{fig:sequential_unlearning_cifar10} 
show the results in the same fashion as seen in Figure \ref{fig:sequential_unlearning_cifar100} for CIFAR-10 and SVHN datasets, and
Table \ref{tab:seq_metrics} 
presents detailed numerical ToW and ToW-MIA results of VTs architectures for five steps.
Here we observe an even higher stability in results with even more negligible degradation, further accentuated by the smaller performance gap between ToW and ToW-MIA than those seen for CIFAR-100. This, along with higher average scores, comes as no surprise due to the higher ``Original'' baseline performance exhibited by the VTs on CIFAR-10.

To further stress-test continual unlearning, we then extend this to 10 sequential unlearning steps under the same setting (NegGrad+ with HR), using both ViT-Small and Swin-Tiny. Results are presented in Figure \ref{fig:sequential_unlearning_10steps} and Table \ref{tab:cu-10steps}, and no substantial accumulated degradation is observed in either ToW or ToW-MIA.

Together, this evidence further supports the observation
from Section \ref{subsec:stab_prox}: 
\textit{Continual unlearning in VTs exhibits high stability across steps, with minimal degradation in performance.}

\begin{table*}[htb]
  \caption{Detailed unlearning performance at each of the 5 sequential steps for the NegGrad+ and Holdout Retraining configuration discussed in Section \ref{subsec:stab_prox} and displayed in Figures \ref{fig:sequential_unlearning_cifar100} and \ref{fig:sequential_unlearning_cifar10}.}
  \centering
  \resizebox{\textwidth}{!}{%
    \begin{tabular}{c c c c c c c c c}
      \toprule
      & \multicolumn{4}{c}{\textbf{CIFAR-10}} & \multicolumn{4}{c}{\textbf{CIFAR-100}} \\
      \cmidrule(lr){2-5} \cmidrule(lr){6-9}
       & \multicolumn{2}{c}{\textbf{ViT-Small}} & \multicolumn{2}{c}{\textbf{Swin-Tiny}} & \multicolumn{2}{c}{\textbf{ViT-Small}} & \multicolumn{2}{c}{\textbf{Swin-Tiny}} \\
      \cmidrule(lr){2-3} \cmidrule(lr){4-5} \cmidrule(lr){6-7} \cmidrule(lr){8-9}
              \textbf{Step}    &\textbf{ToW $(\uparrow)$} & \textbf{ToW-MIA $(\uparrow)$} &\textbf{ToW $(\uparrow)$} & \textbf{ToW-MIA $(\uparrow)$} &\textbf{ToW $(\uparrow)$} & \textbf{ToW-MIA $(\uparrow)$} &\textbf{ToW $(\uparrow)$} & \textbf{ToW-MIA $(\uparrow)$} \\
      \midrule
        1 & $0.907\pm0.030$ & $0.960\pm0.017$ & $0.966\pm0.016$ & $0.933\pm0.082$ & $0.891\pm0.065$ & $0.816\pm0.051$ & $0.906\pm0.039$ & $0.823\pm0.023$ \\
        2 & $0.936\pm0.063$ & $0.944\pm0.029$ & $0.957\pm0.016$ & $0.917\pm0.030$ & $0.868\pm0.031$ & $0.770\pm0.015$ & $0.840\pm0.064$ & $0.764\pm0.085$ \\
        3 & $0.929\pm0.044$ & $0.948\pm0.023$ & $0.954\pm0.054$ & $0.909\pm0.096$ & $0.857\pm0.049$ & $0.764\pm0.047$ & $0.853\pm0.052$ & $0.776\pm0.055$ \\
        4 & $0.943\pm0.010$ & $0.942\pm0.028$ & $0.947\pm0.058$ & $0.899\pm0.058$ & $0.858\pm0.053$ & $0.760\pm0.040$ & $0.843\pm0.021$ & $0.746\pm0.035$ \\
        5 & $0.936\pm0.047$ & $0.936\pm0.031$ & $0.944\pm0.057$ & $0.904\pm0.080$ & $0.851\pm0.006$ & $0.754\pm0.056$ & $0.870\pm0.070$ & $0.781\pm0.089$ \\
      \bottomrule
    \end{tabular}%
  }
  \label{tab:seq_metrics}
\end{table*}

\begin{table}[htb]
\centering
\caption{Continual unlearning over 10 sequential steps on CIFAR-100 using NegGrad+ with the Holdout Retraining proxy (configuration discussed in Section \ref{subsec:stab_prox}).
Results are reported as mean $\pm$ 95\% confidence interval.}
\label{tab:cu-10steps}
\begin{tabular}{lcccc}
\toprule
Step 
& \multicolumn{2}{c}{ViT-Small} 
& \multicolumn{2}{c}{Swin-Tiny} \\
\cmidrule(lr){2-3} \cmidrule(lr){4-5}
& ToW ($\uparrow$) 
& ToW-MIA ($\uparrow$) 
& ToW ($\uparrow$) 
& ToW-MIA ($\uparrow$) \\
\midrule
 1  & $0.947 \pm 0.043$ & $0.889 \pm 0.108$ & $0.947 \pm 0.035$ & $0.860 \pm 0.082$ \\
 2  & $0.971 \pm 0.033$ & $0.828 \pm 0.040$ & $0.900 \pm 0.104$ & $0.786 \pm 0.064$ \\
 3  & $0.973 \pm 0.019$ & $0.845 \pm 0.034$ & $0.953 \pm 0.021$ & $0.850 \pm 0.024$ \\
 4  & $0.918 \pm 0.053$ & $0.840 \pm 0.244$ & $0.940 \pm 0.081$ & $0.816 \pm 0.094$ \\
 5  & $0.921 \pm 0.046$ & $0.769 \pm 0.020$ & $0.958 \pm 0.079$ & $0.834 \pm 0.138$ \\
 6  & $0.966 \pm 0.029$ & $0.807 \pm 0.129$ & $0.921 \pm 0.059$ & $0.817 \pm 0.025$ \\
 7  & $0.951 \pm 0.040$ & $0.841 \pm 0.108$ & $0.908 \pm 0.062$ & $0.792 \pm 0.111$ \\
 8  & $0.961 \pm 0.024$ & $0.817 \pm 0.046$ & $0.939 \pm 0.041$ & $0.798 \pm 0.126$ \\
 9  & $0.946 \pm 0.049$ & $0.807 \pm 0.068$ & $0.936 \pm 0.054$ & $0.822 \pm 0.089$ \\
 10 & $0.923 \pm 0.080$ & $0.825 \pm 0.115$ & $0.921 \pm 0.072$ & $0.803 \pm 0.100$ \\
\bottomrule
\end{tabular}
\end{table}

\subsection{Detailed ToW and ToW-MIA Results 
}
This section presents detailed ToW and ToW-MIA performance results for all evaluated architectures on the CIFAR-10, CIFAR-100, and SVHN datasets. 
We first present results for CIFAR-10 and CIFAR-100 datasets, for ViT-Small, Swin-Tiny (Table \ref{tab:all_unlearn_metrics}), and ResNets (Table \ref{tab:cnn_unlearn_metrics}), which are used to generate Figures \ref{fig:cifar100_comparison} and \ref{fig:cifar10_unlearning_comparison}.

\subsubsection{ViT, Swin-T, and DINOv2 Results Across All Algorithms}
Table \ref{tab:all_unlearn_metrics} provides detailed ToW and ToW-MIA results for ViT-Small, Swin-Tiny and DINOv2-Small on CIFAR-10/CIFAR-100. These values are used in the corresponding architecture and method comparison figures.
\label{app:perfVTscifars}
\begin{table}[htb]
  \caption{Detailed performance results for the VTs (ViT, Swin-T, DINOv2) across both the CIFAR datasets.  
  }
  \centering
  \begin{subtable}[t]{\columnwidth}
    \centering
    \begin{tabular}{ccc@{\hspace{1em}}cc}
      \toprule
      & \multicolumn{2}{c}{\textbf{Confidence}} & \multicolumn{2}{c}{\textbf{Holdout Retraining}} \\
      \cmidrule(lr){2-3} \cmidrule(lr){4-5}
      \textbf{Algorithm} & \textbf{ToW $(\uparrow)$} & \textbf{ToW-MIA $(\uparrow)$} &\textbf{ToW $(\uparrow)$} & \textbf{ToW-MIA $(\uparrow)$} \\
      \midrule
      Original  & $0.940\pm0.005$ & $0.895\pm0.007$          & $0.891\pm0.016$  & $0.831\pm0.014$  \\
       Fine-tune \textbf{  } & $0.934\pm0.022$ & $0.919\pm0.040$          & $0.928\pm0.013$  & $0.913\pm0.008$  \\
      NegGrad+ \textbf{  }  & $0.969\pm0.013$ & $0.938\pm0.015$          & $0.916\pm0.016$  & $0.968\pm0.015$  \\
      SalUn \textbf{  }     & $0.975\pm0.003$ & $0.582\pm0.049$          & $0.956\pm0.003$  & $0.766\pm0.039$  \\
      \bottomrule
    \end{tabular}
    \caption{CIFAR-10 -- ViT-Small}
  \end{subtable}
   
  \begin{subtable}[t]{\columnwidth}
    \centering
    \begin{tabular}{ccc@{\hspace{1em}}cc}
      \toprule
      & \multicolumn{2}{c}{\textbf{Confidence}} & \multicolumn{2}{c}{\textbf{Holdout Retraining}} \\
      \cmidrule(lr){2-3} \cmidrule(lr){4-5}
      \textbf{Algorithm} &\textbf{ToW $(\uparrow)$} & \textbf{ToW-MIA $(\uparrow)$} &\textbf{ToW $(\uparrow)$} & \textbf{ToW-MIA $(\uparrow)$} \\
      \midrule
      Original  & $0.915\pm0.027$ & $0.866\pm0.014$          & $0.862\pm0.008$  & $0.811\pm0.018$  \\
       Fine-tune \textbf{  } & $0.930\pm0.019$ & $0.922\pm0.015$          & $0.923\pm0.021$  & $0.931\pm0.021$  \\
      NegGrad+ \textbf{  }  & $0.971\pm0.012$ & $0.922\pm0.019$          & $0.977\pm0.023$  & $0.924\pm0.028$  \\
      SalUn \textbf{  }     & $0.963\pm0.048$ & $0.723\pm0.116$          & $0.961\pm0.048$  & $0.834\pm0.022$  \\
      \bottomrule
    \end{tabular}
    \caption{CIFAR-10 -- Swin-Tiny}
  \end{subtable}

  \begin{subtable}[t]{\columnwidth}
    \centering
    \begin{tabular}{ccc@{\hspace{1em}}cc}
      \toprule
      & \multicolumn{2}{c}{\textbf{Confidence}} & \multicolumn{2}{c}{\textbf{Holdout Retraining}} \\
      \cmidrule(lr){2-3} \cmidrule(lr){4-5}
      \textbf{Algorithm} &\textbf{ToW $(\uparrow)$} & \textbf{ToW-MIA $(\uparrow)$} &\textbf{ToW $(\uparrow)$} & \textbf{ToW-MIA $(\uparrow)$} \\
      \midrule
      Original  & $0.629\pm0.038$ & $0.521\pm0.026$          & $0.619\pm0.040$  & $0.538\pm0.023$  \\
       Fine-tune \textbf{  } & $0.813\pm0.008$ & $0.831\pm0.014$          & $0.855\pm0.036$  & $0.889\pm0.008$  \\
      NegGrad+ \textbf{  }  & $0.844\pm0.034$ & $0.736\pm0.023$          & $0.931\pm0.039$  & $0.838\pm0.024$  \\
      SalUn \textbf{  }     & $0.923\pm0.019$ & $0.602\pm0.014$          & $0.903\pm0.049$  & $0.709\pm0.085$  \\
      \bottomrule
    \end{tabular}
    \caption{CIFAR-100 -- ViT-Small}
  \end{subtable}
  
  \begin{subtable}[t]{\columnwidth}
    \centering
    \begin{tabular}{ccc@{\hspace{1em}}cc}
      \toprule
      & \multicolumn{2}{c}{\textbf{Confidence}} & \multicolumn{2}{c}{\textbf{Holdout Retraining}} \\
      \cmidrule(lr){2-3} \cmidrule(lr){4-5}
      \textbf{Algorithm} &\textbf{ToW $(\uparrow)$} & \textbf{ToW-MIA $(\uparrow)$} &\textbf{ToW $(\uparrow)$} & \textbf{ToW-MIA $(\uparrow)$} \\
      \midrule
      Original  & $0.665\pm0.014$ & $0.553\pm0.015$          & $0.670\pm0.027$  & $0.586\pm0.025$  \\
      Fine-tune \textbf{  } & $0.774\pm0.030$ & $0.758\pm0.015$          & $0.822\pm0.034$  & $0.839\pm0.032$  \\
      NegGrad+ \textbf{  }  & $0.867\pm0.021$ & $0.767\pm0.019$          & $0.975\pm0.027$  & $0.902\pm0.035$  \\
      SalUn \textbf{  }     & $0.835\pm0.060$ & $0.557\pm0.034$          & $0.870\pm0.039$  & $0.751\pm0.043$  \\
      \bottomrule
    \end{tabular}
     \caption{CIFAR-100 -- Swin-Tiny}
  \end{subtable}
\begin{subtable}[t]{\columnwidth}
  \centering
  \begin{tabular}{ccc@{\hspace{1em}}cc}
    \toprule
    & \multicolumn{2}{c}{\textbf{Confidence}} & \multicolumn{2}{c}{\textbf{Holdout Retraining}} \\
    \cmidrule(lr){2-3} \cmidrule(lr){4-5}
    \textbf{Algorithm} & \textbf{ToW $(\uparrow)$} & \textbf{ToW-MIA $(\uparrow)$} & \textbf{ToW $(\uparrow)$} & \textbf{ToW-MIA $(\uparrow)$} \\
    \midrule
    Original   & $0.578\pm0.045$ & $0.455\pm0.045$ & $0.555\pm0.112$ & $0.425\pm0.080$ \\
    Fine-tune  & $0.867\pm0.009$ & $0.802\pm0.042$ & $0.846\pm0.038$ & $0.824\pm0.070$ \\
    NegGrad+   & $0.878\pm0.041$ & $0.786\pm0.064$ & $0.937\pm0.044$ & $0.896\pm0.032$ \\
    SalUn      & $0.747\pm0.105$ & $0.701\pm0.042$ & $0.831\pm0.121$ & $0.840\pm0.125$ \\
    \bottomrule
  \end{tabular}
  \caption{CIFAR-100 -- DINOv2-Small}
\end{subtable}

  \label{tab:all_unlearn_metrics}
\end{table}

\subsubsection{ResNet-18 and ResNet-50 Results Across All Algorithms}
Table \ref{tab:cnn_unlearn_metrics} reports the detailed results for ResNet-18 (CIFAR-10) and ResNet-50 (CIFAR-100). These metrics support the comparative analysis shown in Figures 
\ref{fig:cifar100_comparison} and \ref{fig:cifar10_unlearning_comparison}.
Further breakdowns of ToW and ToW-MIA (by individual terms) for Vision Transformer architectures across datasets are reported in Table \ref{tab:perform_met}.

\begin{table}[htb]
  \caption{Detailed performance result metrics for ResNet-18 and ResNet-50 on the CIFAR-10 and CIFAR-100 datasets respectively. 
  }
  \centering
  \begin{subtable}[t]{\columnwidth}
    \centering
    \begin{tabular}{ccc@{\hspace{1em}}cc}
      \toprule
      & \multicolumn{2}{c}{\textbf{Confidence}} & \multicolumn{2}{c}{\textbf{Holdout Retraining}} \\
      \cmidrule(lr){2-3} \cmidrule(lr){4-5}
      \textbf{Algorithm} & \textbf{ToW $(\uparrow)$} & \textbf{ToW-MIA $(\uparrow)$} &\textbf{ToW $(\uparrow)$} & \textbf{ToW-MIA $(\uparrow)$} \\
      \midrule
      Fine-tune  \textbf{ \textbf{  }} & $0.919 \pm 0.042$ & $0.920 \pm 0.059$ & $0.920 \pm 0.035$ & $0.865 \pm 0.033$ \\
      NegGrad+  \textbf{  } & $0.880 \pm 0.039$ & $0.843 \pm 0.074$ & $0.870 \pm 0.033$ & $0.791 \pm 0.040$ \\
      SalUn  \textbf{  } & $0.859 \pm 0.042$ & $0.867 \pm 0.092$ & $0.861 \pm 0.044$ & $0.783 \pm 0.028$ \\
      \bottomrule
    \end{tabular}
    \caption{CIFAR-10 with ResNet-18}
  \end{subtable}

  \begin{subtable}[t]{\columnwidth}
    \centering
    \begin{tabular}{ccc@{\hspace{1em}}cc}
      \toprule
      & \multicolumn{2}{c}{\textbf{Confidence}} & \multicolumn{2}{c}{\textbf{Holdout Retraining}} \\
      \cmidrule(lr){2-3} \cmidrule(lr){4-5}
      \textbf{Algorithm} & \textbf{ToW $(\uparrow)$} & \textbf{ToW-MIA $(\uparrow)$} &\textbf{ToW $(\uparrow)$} & \textbf{ToW-MIA $(\uparrow)$} \\
      \midrule
      Fine-tune  \textbf{  } & $0.863 \pm 0.049$ & $0.857 \pm 0.059$ & $0.846 \pm 0.032$ & $0.792 \pm 0.044$ \\
      NegGrad+  \textbf{  } & $0.890 \pm 0.047$ & $0.922 \pm 0.017$ & $0.866 \pm 0.031$ & $0.838 \pm 0.027$ \\
      SalUn  \textbf{  } & $0.665 \pm 0.031$ & $0.636 \pm 0.038$ & $0.696 \pm 0.032$ & $0.699 \pm 0.034$ \\
      \bottomrule
    \end{tabular}
    \caption{CIFAR-100 with ResNet-50}
  \end{subtable}
  \label{tab:cnn_unlearn_metrics}
\end{table}


\begin{table*}[htb]
  \caption{Accuracies on $D_r$, $D_f$ and $D_{test}$ for the ``Retrain'' models $\theta_r$  as well as the respective ``Unlearned'' models $\theta_u$  , evaluated across all unlearning algorithms within the RUM$_F$ framework.
  MIA performance is also reported. 
  Results are averaged over 3 runs, with 95\% confidence intervals.}
  \centering
  
  \begin{subtable}[t]{\textwidth}
    \centering
    \resizebox{\textwidth}{!}{%
      \begin{tabular}{l cccc @{\hspace{1em}} cccc}
        \toprule
         & \multicolumn{4}{c}{\textbf{Confidence}} & \multicolumn{4}{c}{\textbf{Holdout Retraining}} \\
        \cmidrule(lr){2-5} \cmidrule(lr){6-9}
        \textbf{Algorithm} & \textbf{Retain Acc} & \textbf{Forget Acc} & \textbf{Test Acc} & \textbf{MIA} & \textbf{Retain Acc} & \textbf{Forget Acc} & \textbf{Test Acc} & \textbf{MIA} \\
        \midrule
        Retrain   &  $99.969\pm0.006$ & $94.089\pm0.456$ & $93.703\pm0.486$ & $0.108\pm0.007$ & $99.974\pm0.035$ & $89.244\pm1.321$ & $93.767\pm0.029$ & $0.174\pm0.012$ \\
        Fine-tune & $99.433\pm0.639$ & $94.111\pm0.912$ & $87.947\pm1.519$ & $0.128\pm0.014$ & $99.562\pm0.347$ & $87.889\pm2.356$ & $88.240\pm0.348$ & $0.204\pm0.016$ \\
        NegGrad+ & $99.832\pm0.140$ & $93.411\pm1.078$ & $91.407\pm0.872$ & $0.069\pm0.008$ & $99.814\pm0.073$ & $83.289\pm2.115$ & $91.340\pm0.413$ & $0.167\pm0.020$ \\
        SalUn & $100.000\pm0.000$ & $96.033\pm0.543$ & $93.203\pm0.617$ & $0.523\pm0.057$ & $100.000\pm0.000$ & $93.089\pm1.971$ & $93.180\pm0.523$ & $0.403\pm0.042$ \\
        \bottomrule
      \end{tabular}%
    }
    \caption{CIFAR-10 with ViT-Small}
  \end{subtable}
  
  \begin{subtable}[t]{\textwidth}
    \centering
    \resizebox{\textwidth}{!}{%
      \begin{tabular}{l cccc @{\hspace{1em}} cccc}
        \toprule
         & \multicolumn{4}{c}{\textbf{Confidence}} & \multicolumn{4}{c}{\textbf{Holdout Retraining}} \\
        \cmidrule(lr){2-5} \cmidrule(lr){6-9}
        \textbf{Algorithm} & \textbf{Retain Acc} & \textbf{Forget Acc} & \textbf{Test Acc} & \textbf{MIA} & \textbf{Retain Acc} & \textbf{Forget Acc} & \textbf{Test Acc} & \textbf{MIA} \\
        \midrule
        Retrain    & $99.961\pm0.021$ & $91.867\pm1.711$ & $91.607\pm0.596$ & $0.133\pm0.002$ & $99.979\pm0.027$ & $86.389\pm1.246$ & $91.767\pm0.689$ & $0.190\pm0.020$ \\
        Fine-tune & $99.833\pm0.141$ & $96.733\pm0.840$ & $89.463\pm0.660$ & $0.076\pm0.005$ & $99.779\pm0.012$ & $91.478\pm0.208$ & $89.183\pm0.352$ & $0.148\pm0.009$ \\
        NegGrad+ & $99.968\pm0.018$ & $93.833\pm0.597$ & $90.643\pm1.096$ & $0.064\pm0.005$ & $99.903\pm0.024$ & $81.911\pm0.694$ & $90.473\pm0.254$ & $0.181\pm0.007$ \\
        SalUn & $100.000\pm0.000$ & $93.967\pm8.103$ & $90.743\pm0.547$ & $0.403\pm0.126$ & $100.000\pm0.000$ & $88.533\pm9.399$ & $91.160\pm0.692$ & $0.351\pm0.043$ \\
        \bottomrule
      \end{tabular}%
    }
    \caption{CIFAR-10 with Swin-Tiny}
  \end{subtable}
  
  \begin{subtable}[t]{\textwidth}
    \centering
    \resizebox{\textwidth}{!}{%
      \begin{tabular}{l cccc @{\hspace{1em}} cccc}
        \toprule
         & \multicolumn{4}{c}{\textbf{Confidence}} & \multicolumn{4}{c}{\textbf{Holdout Retraining}} \\
        \cmidrule(lr){2-5} \cmidrule(lr){6-9}
        \textbf{Algorithm} & \textbf{Retain Acc} & \textbf{Forget Acc} & \textbf{Test Acc} & \textbf{MIA} & \textbf{Retain Acc} & \textbf{Forget Acc} & \textbf{Test Acc} & \textbf{MIA} \\
        \midrule
        Retrain    & $98.457\pm0.819$ & $69.322\pm1.788$ & $69.670\pm2.445$ & $0.432\pm0.012$ & $98.063\pm0.413$ & $68.767\pm3.828$ & $69.467\pm1.032$ & $0.412\pm0.024$. \\
        Fine-tune & $99.325\pm0.148$ & $86.056\pm1.907$ & $68.193\pm1.578$ & $0.283\pm0.014$ & $99.322\pm0.622$ & $81.200\pm2.766$ & $68.350\pm1.433$ & $0.322\pm0.025$ \\
        NegGrad+  & $99.846\pm0.166$ & $82.200\pm0.299$ & $71.440\pm1.793$ & $0.192\pm0.004$ & $99.637\pm0.177$ & $72.722\pm0.621$ & $70.997\pm0.448$ & $0.276\pm0.006$ \\
        SalUn     & $99.982\pm0.007$ & $64.756\pm0.669$ & $71.467\pm0.816$ & $0.810\pm0.014$ & $99.981\pm0.006$ & $63.878\pm2.161$ & $72.620\pm0.480$ & $0.666\pm0.068$ \\
        \bottomrule
      \end{tabular}%
    }
    \caption{CIFAR-100 with ViT-Small}
  \end{subtable}
  
  \begin{subtable}[t]{\textwidth}
    \centering
    \resizebox{\textwidth}{!}{%
      \begin{tabular}{l cccc @{\hspace{1em}} cccc}
        \toprule
         & \multicolumn{4}{c}{\textbf{Confidence}} & \multicolumn{4}{c}{\textbf{Holdout Retaining}} \\
        \cmidrule(lr){2-5} \cmidrule(lr){6-9}
        \textbf{Algorithm} & \textbf{Retain Acc} & \textbf{Forget Acc} & \textbf{Test Acc} & \textbf{MIA} & \textbf{Retain Acc} & \textbf{Forget Acc} & \textbf{Test Acc} & \textbf{MIA} \\
        \midrule
        Retrain     & $99.220\pm0.167$ & $69.833\pm1.242$ & $68.923\pm0.296$ & $0.430\pm0.015$ & $99.143\pm0.185$ & $70.267\pm2.848$ & $69.057\pm0.827$ & $0.392\pm0.024$ \\
        Fine-tune & $99.638\pm0.184$ & $91.811\pm3.019$ & $68.503\pm0.405$ & $0.194\pm0.022$ & $99.368\pm0.557$ & $87.289\pm1.300$ & $68.377\pm1.536$ & $0.239\pm0.016$ \\
        NegGrad+ & $99.940\pm0.047$ & $81.900\pm0.955$ & $69.410\pm1.083$ & $0.207\pm0.002$ & $99.914\pm0.047$ & $70.111\pm0.723$ & $69.833\pm0.521$ & $0.308\pm0.009$ \\
        SalUn & $99.965\pm0.003$ & $55.578\pm6.181$ & $70.750\pm0.676$ & $0.859\pm0.054$ & $99.966\pm0.007$ & $60.156\pm1.615$ & $71.490\pm0.696$ & $0.616\pm0.021$ \\
        \bottomrule
      \end{tabular}%
    }
    \caption{CIFAR-100 with Swin-Tiny}
  \end{subtable}
  \label{tab:perform_met}
\end{table*}


\end{document}